%% file: main.tex
\documentclass[11pt]{article}

\usepackage[final]{acl}

\usepackage{times}
\usepackage[utf8]{inputenc}%
\usepackage[T1]{fontenc}%
\usepackage{hyperref}%
\usepackage{booktabs}%
\usepackage{amsfonts}%
\usepackage{microtype}%
\usepackage{xcolor}%
\usepackage{colortbl}%
\usepackage{graphicx}
\usepackage{multirow}
\usepackage[most]{tcolorbox}
\usepackage{cleveref}
\usepackage{threeparttable}
\usepackage{makecell}%
\usepackage{listings}%
\usepackage{dblfloatfix}%
\usepackage{placeins}%

\newtcolorbox{takeaway}{
breakable,
  colback=gray!8,
  colframe=gray!45,
  boxrule=0.4pt,
  arc=1pt,
  left=5pt,
  right=5pt,
  top=5pt,
  bottom=5pt,
  before skip=6pt,
  after skip=6pt
}

\title{Assessing Adversarial Robustness of Latent Reasoning Models}

\author{
  Shaolong Chen\textsuperscript{1}\thanks{Equal Contribution.}\quad
  Ang Li\textsuperscript{2}\footnotemark[1]\quad
  Mingjie Li\textsuperscript{3}\quad
  Yisen Wang\textsuperscript{2}\thanks{\raggedright Corresponding author: Yisen Wang (\href{mailto:yisen.wang@pku.edu.cn}{yisen.wang@pku.edu.cn}).} \\
   \textsuperscript{1}Yuanpei College, Peking University \\
   \textsuperscript{2}State Key Lab of General Artificial Intelligence,\\ School of Intelligence Science and Technology, Peking University \\
   \textsuperscript{3}CISPA Helmholtz Center for Information Security 
}

\begin{document}

\maketitle

\begin{abstract}
  \input{content/abstract}
\end{abstract}

\section{Introduction}
\input{content/introduction}

\section{Related Works}
\input{content/related_works}

\section{Methodology}

\begin{figure*}[htbp]
    \centering
    \includegraphics[width=1\linewidth]{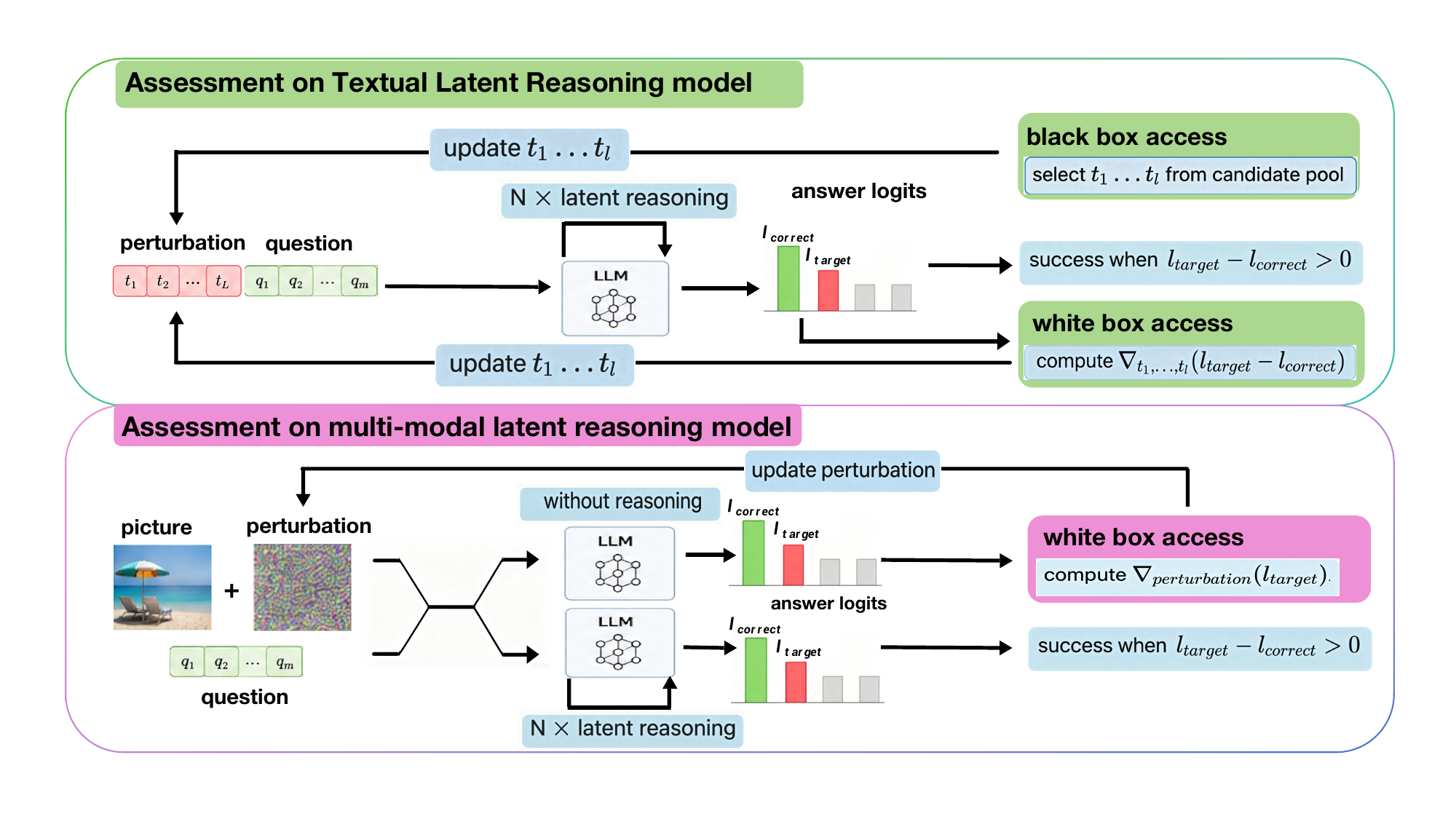}
    \caption{Assessment protocol of our paper: the figure shows perturbation generation for textual models via black-box search or white-box optimization, and illustrates white-box adversarial image perturbation for multi-modal models.}
    \label{fig:placeholder}
\end{figure*}

\input{content/method}

\input{content/experiments}

\section{Discussion and Conclusion}
\input{content/conclusion}

\clearpage

\section*{Limitations}
\input{content/limitation}

\section*{Acknowledgments}

Yisen Wang is supported by National Natural Science Foundation of China (62376010, 92370129), Beijing Major Science and Technology Project under Contract no. Z251100008425006, Beijing Natural Science Foundation (No. L257007), Beijing Nova Program (20230484344, 20240484642), and State Key Laboratory of General Artificial Intelligence.

\bibliography{main}

\clearpage

\appendix

\input{content/appendix}

\end{document}

%% file: content/abstract.tex
Large language models increasingly rely on long chain-of-thought (CoT) trajectories for complex reasoning, but autoregressive generation brings substantial memory and inference costs. Latent reasoning models (LRMs) offer a more efficient alternative by compressing intermediate reasoning into a small number of continuous latent vectors. Despite their efficiency, however, the adversarial robustness of LRMs remains largely underexplored. In this work, we systematically evaluate the robustness of latent reasoning across textual and multimodal settings, covering eight models and six benchmarks. We find that, across our evaluated settings, LRMs are generally less robust than explicit CoT baselines under adversarial perturbations, with particularly severe degradation under white-box attacks. Further analysis reveals distinct failure modes across modalities: textual latent states exhibit brittle dynamics and high sensitivity to specific input patterns, while latent states in multimodal models can remain largely invariant to input perturbations and have limited influence on final predictions. These findings expose robustness limitations of current latent reasoning approaches and highlight the need to jointly consider efficiency and robustness when designing implicit reasoning systems. We have open-sourced our code to facilitate reproduction of our research \url{https://github.com/PKU-ML/latent-reasoning-model-assessment}.

%% file: content/introduction.tex
Large language models (LLMs) have demonstrated strong performance in domains such as mathematics, coding, and agentic tasks \cite{luo2025large, jiang2026codegenerationsurvey, xu2025towardreasoningsurvey}. 
These achievements are driven by two complementary forces: scaling pre-training compute to store more world knowledge in model parameters~\cite{kaplan2020scaling}, and scaling inference-time compute to support more deliberative reasoning and exploration in context~\cite{chain_of_thought,o1,zhang2025survey}.
The latter typically relies on auto-regressively generated discrete reasoning tokens, and has achieved notable success with high-quality chain-of-thought data \cite{guha2026openthoughts} and large-scale reinforcement learning \cite{guo2025deepseek}. 
However, such explicit reasoning still faces fundamental constraints imposed by the sequential nature of auto-regressive generation, finite context windows, and the limited information capacity of individual tokens, raising concerns about the trade-off between effectiveness and efficiency \cite{gao2026far}.

Latent space reasoning, where intermediate thoughts are stored and updated in the form of continuous representations, has been proposed as an alternative to reasoning with discrete tokens \cite{coconut, monet, chen2025reasoning}. 
By avoiding artificial linguistic constraints and the tokenization bottleneck, latent reasoning has shown the potential to rival discrete reasoning in accuracy with only several forward passes \cite{coconut,codi}.  
However, continuous reasoning chains may introduce a new robustness concern. 
Unlike discrete token sequences, which are difficult to perturb imperceptibly, continuous vectors are known to be sensitive to small adversarial changes \cite{goodfellow2014explaining, pgd}. 
Given that discrete LLMs are already vulnerable to adversarial inputs in both safety-critical settings \cite{gcg} and reasoning tasks \cite{gan2024reasoning, ho2026format}, we hypothesize that reasoning in continuous latent spaces may create a larger surface for subtle but influential perturbations, leading to weaker resilience against input perturbations. This motivates our central question:

\begin{center}
    \textit{How robust are Latent Reasoning Models against input perturbations?}
\end{center}

\begin{figure}[t]
    \centering
    \includegraphics[width=1\linewidth]{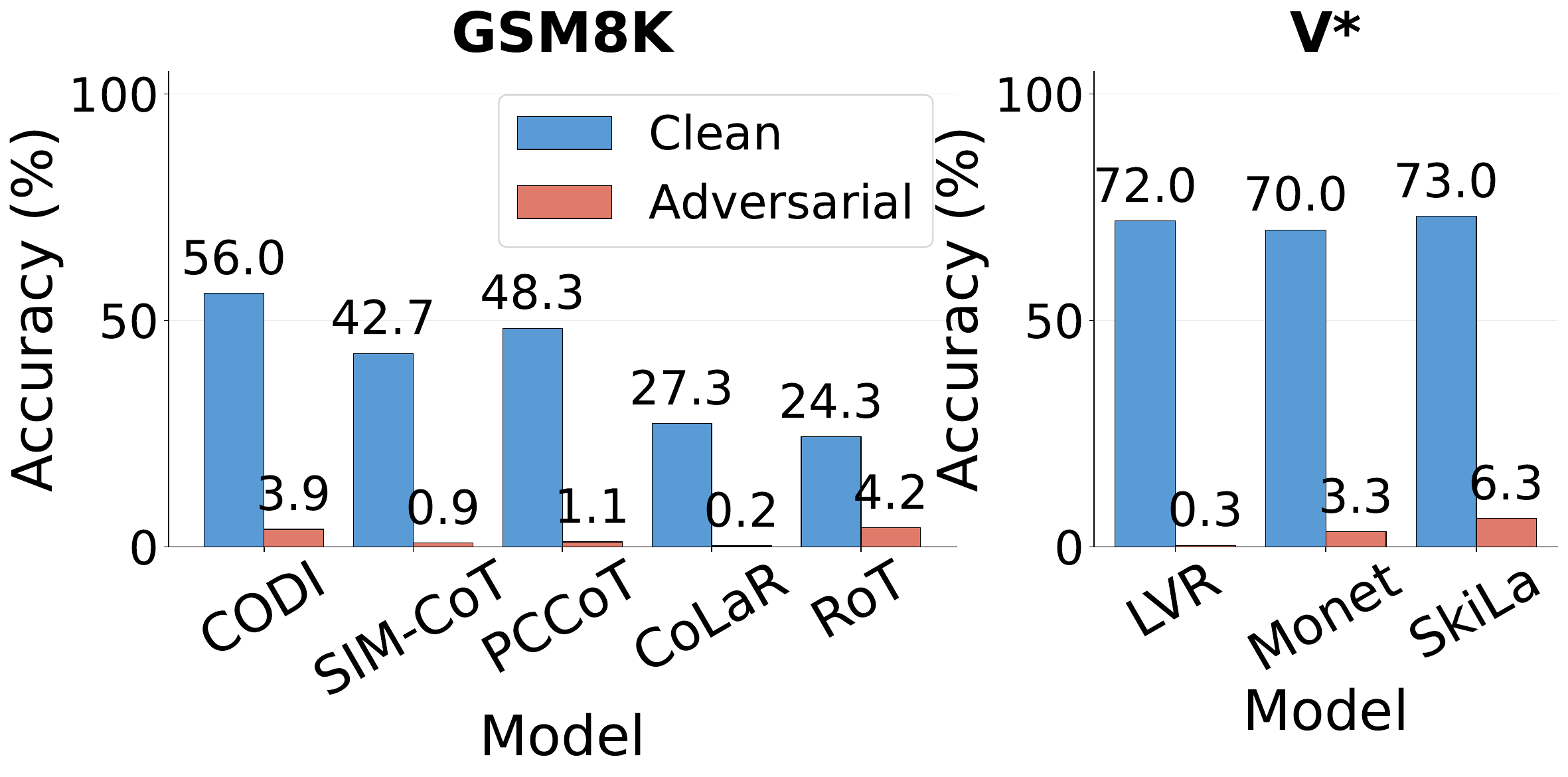}
    \caption{Evaluation of latent reasoning models on the GSM8K (left) and V$^*$ (right) datasets, comparing accuracy under clean (blue bars) and adversarial (red bars) conditions.}
    \label{fig:intro}
\end{figure}

In this work, we study this question through adversarial inputs, i.e., problems carefully modified to mislead the model to wrong answer while remaining semantically and logically equivalent to the original ones. Technically, to craft such inputs, we adapt existing gradient-based and heuristic-based searching methods previously designed for discrete reasoning LLMs to the latent reasoning models. 
To examine whether the observed trends hold across modalities and model families, our empirical study includes both single-modal (textual) and multimodal (visual)
latent reasoning models, spanning 8 representative LRMs and 6 benchmarks. 

Our study reveals both similarities and differences between discrete and latent reasoning models in terms of adversarial robustness. Similar to explicit reasoning models, we observe that latent reasoning models are consistently prone to adversarial inputs (\cref{sec:text_model_results} and \cref{sec:multi_modal_model_results}), with \cref{fig:intro} showing that all of the 8 models exhibits at least 80\% of relative performance drop under adversarial settings. Beyond similarities, we find that textual latent reasoning models are more vulnerable under white-box perturbations and heuristic black-box queries than the explicit reasoning baseline in our evaluated settings (\cref{sec:text_model_results}). 
Moreover, to understand the mechanism behind our attack, we show that adversarial perturbations substantially reduce the cosine similarity of latent tokens in textual reasoning models (\cref{sec:similarity_results}). Finally, we find that latent tokens in mult-modal models remain comparatively unchanged under perturbations, and our ablation study on clean and adversarial inputs further shows that the visual latent tokens can be functionally redundant, exposing a core limitation of existing works (\cref{sec:reasoning_results}).

To summarize, our contributions are threefold:
\begin{itemize}
    \item We address the underexplored adversarial robustness of implicit reasoning models through a systematic evaluation spanning textual and visual modalities, 8 models, and 6 datasets.
    \item As far as our study concerns, current implicit reasoners are generally less robust than explicit chain-of-thought baselines under adversarial perturbations, especially under white-box setting.
    \item Our further investigation attribute this gap to the 
    brittle latent dynamics and unused latent states that bypass reasoning.
\end{itemize}

%% file: content/related_works.tex
Our work connects two lines of research: latent reasoning and robustness evaluation. We first review latent reasoning methods and then summarize robustness studies of language and vision-language reasoning models. In this paper, we use the terms implicit reasoning and latent reasoning interchangeably to refer to this form of continuous internal reasoning.

\subsection{Latent Reasoning in LLMs}

Large language models generate reasoning steps in discrete tokens, typically in the form of chain-of-thought rationales~\cite{chain_of_thought}. Although token-by-token reasoning improves performance by decomposing problems into more approachable parts, its latency increases linearly as length grows and memory cost increases quadratically, which becomes expensive for long trajectories that easily reaches tens of thousands of tokens  \cite{sui2025stop, cuadron2025danger}.

Latent-reasoning methods instead encode intermediate computation in continuous hidden states, with the aim of reducing the decoding overhead and discretization imposed by textual rationales.  Existing methods typically reason with only around 10 latent tokens, substantially shortening the reasoning trajectory and reducing inference cost. Early work focused on textual reasoning, with \citet{coconut} enabling parallel search in latent space through recurrent feedback of hidden states and \citet{codi} distilling explicit reasoning chains into continuous representations. \citet{colar} realizes dynamic-speed reasoning through next compressed embedding prediction. Recently, this paradigm has extended to visual reasoning, where \citet{lvr} and \citet{monet} equip vision-language models with the capacity for internal visual manipulation and multi-step reasoning within latent space.

\subsection{Assessing Robustness of LLMs Reasoning}

Textual and Multimodal Large Language Models (MLLMs) both exhibit notable fragility under subtle input perturbations: small changes to textual prompts or visual inputs can lead to substantially different model outputs~\cite{llm_pertur,vlm_pertur}, motivating robustness assessments of both types of models.

For textual reasoning models, \citet{seca} investigate the robustness of LLMs under semantically equivalent and coherent prompt rewrites, demonstrating that even meaning-preserving reformulations can lead to substantial performance decrease. In addition, \citet{prompt_var} assesses the robustness of LLMs against prompt variations with irrelevant context and other input perturbations, revealing the vulnerability of current models to distracting or noisy contexts. For multimodal large language models (MLLMs), prior studies such as \citet{vlm_adv} and \citet{vlm_robust} mainly focus on assessing robustness against perturbations applied to the input images.

The studies above primarily assess the robustness of models that generate explicit reasoning traces. More recently, a few studies have begun to examine the robustness of latent reasoning models. \citet{text} use internal perturbations and token swapping to study the causal role of latent tokens in textual models, while \citet{mm} analyzes the stability and functional role of latent tokens in multi-modal models. Unlike these studies, which primarily diagnose latent representations through internal interventions, we assess model-level robustness to task-preserving adversarial inputs across both textual and multi-modal models, including white-box and black-box settings and comparisons with explicit reasoning models.

%% file: content/method.tex
To assess the robustness of latent reasoning, we are required to perturb the input questions in a way that misleads LRMs to wrong outputs while preserving the underlying answer unchanged. Our main methods are inspired by existing techniques for crafting adversarial inputs against LLMs and MLLMs, which we provide detailed introduction below.

\subsection{Preliminary}
\paragraph{Latent Reasoning.}
As discussed in Related Works, latent reasoning models span both language and vision-language modalities and can be trained with diverse objectives and supervision signals. Despite these differences, mainstream latent reasoning models generally follow a similar inference paradigm:

Given an input sequence,
\[
x = (x_1, x_2, \dots, x_T),
\]
the model first generates a sequence of latent hidden states:
\[
h_i = f_\theta(x_{\le T}, h_{<i}),
\]
where $h_i$ denotes the $i$-th latent reasoning token, represented as a continuous hidden-state vector rather than a discrete text token,  which may encode aspects of the reasoning process, such as intermediate computation results.

After completing latent reasoning, the model outputs the final answer:
\[
y_i = f_\theta(x_{\le T}, h_{\le N}, y_{<i}),
\]
where $h_{\le N}$ represents the complete latent reasoning tokens and $y_{<i}$ denotes previously generated output tokens.

\paragraph{Threat Models.}
We assess robustness by introducing input perturbations that preserve the answer and the underlying solution process while causing incorrect predictions, considering two standard threat settings:

\begin{itemize}
\item \textbf{White-box setting.} The adversary has complete transparency into the target model. This allows the adversary to directly optimize perturbations (e.g., adversarial prefixes or images) with gradient-based methods. We use the textual and multi-modal white-box assessments as worst-case stress tests designed to expose model vulnerabilities.

\item \textbf{Black-box setting.} The adversary cannot access the model internals and can only interact with the model through query-response interactions. We evaluate average robustness against a fixed, non-adaptive set of input perturbations.
\end{itemize}

Thus, the two protocols assess robustness from two complementary perspectives: white-box assessments measure worst-case vulnerability, whereas black-box assessments measure average robustness under a fixed perturbation set.

\subsection{White-Box Assessment}

We instantiate the white-box assessment in two settings: textual reasoning and multi-modal reasoning, as detailed below. Throughout this section, $\oplus$ denotes the modality-specific application of a perturbation to an input: prefix concatenation for text and pixel-wise addition for images.

\paragraph{Textual Reasoning.} For each problem $x$, we prepend a short adversarial prefix $x_\mathrm{pre}$ to the original question and optimize the prefix toward a targeted incorrect answer:
\begin{equation}
\max_{x_\mathrm{pre} \in \mathcal{T}^{L}} \left(p_{\mathrm{target}}(x_\mathrm{pre} \oplus x) - p_{\mathrm{correct}}(x_\mathrm{pre} \oplus x) \right),
\end{equation}
where $x_\mathrm{pre} \oplus x$ denotes the perturbed textual input formed by prepending $x_\mathrm{pre}$ to $x$. The functions $p_{\mathrm{target}}(\cdot)$ and $p_{\mathrm{correct}}(\cdot)$ denote the model probabilities assigned to the targeted incorrect answer and the correct answer, respectively.
We optimize this discrete prefix using the GCG method~\cite{gcg}.

At each iteration, we compute gradients of the loss with respect to token embeddings, select candidate substitutions from the top-$K$ tokens with the highest gradient scores, and update the prefix by assessing multiple sampled candidates. This process is repeated until convergence or until the model outputs an incorrect answer.

\paragraph{Multi-Modal Reasoning.} For multi-modal reasoning models with image $I$ and question $x$, we construct a bounded adversarial image perturbation $\delta$ satisfying $\|\delta\|_{\infty} \leq \epsilon$, where $\epsilon = \frac{4}{255}$. We optimize $\delta$ to increase the logit of a targeted incorrect answer token:
\begin{equation}
    \max_{\|\delta\|_{\infty} \leq \epsilon} \ p_{\mathrm{target}}(x; I \oplus \delta),
\end{equation}
where $I \oplus \delta$ denotes the adversarial image formed by applying the pixel-wise perturbation $\delta$ to $I$, and $p_{\mathrm{target}}(x; I \oplus \delta)$ denotes the probability assigned to the targeted incorrect token at the selected position given question $x$ and the adversarial image.
Specifically, we update $\delta$ using PGD~\cite{pgd}:
\begin{equation}
\delta \leftarrow \mathrm{clamp}\left(\delta + \alpha \cdot \mathrm{sign}\left(\nabla_\delta \mathcal{L}(x; I \oplus \delta)\right), -\epsilon, \epsilon\right),
\end{equation}
where $\mathcal{L} = p_{\mathrm{target}}$. We repeat this update until the model produces an incorrect answer.

We optimize $\delta$ using logits at the question's terminating punctuation, immediately before latent reasoning begins. As discussed in \cref{sec:reasoning_results}, the perturbation can alter the model's initial preference, which may in turn influence the final output.

\input{content/text_whitebox_table}

\subsection{Black-Box Assessment}
\label{sec:black_box_assessment}

As current textual LRMs are mainly evaluated on mathematical reasoning problems, we propose a heuristic perturbation strategy against them, which prepends irrelevant contexts containing numbers. Such perturbations preserve the semantics and correct answer while introducing plausible numerical distractions. Without access to model parameters, we assess black-box robustness through query-based interactions using a fixed, non-adaptive pool $\mathcal{C}$ of such contexts.

Specifically, given a question $x$ and its ground-truth answer $a^{*}$, we prepend each irrelevant context $c \in \mathcal{C}$ containing a number to form a rewritten question $x' = c \oplus x$.  Given a test set $\mathcal{D}=\{(x_i,a_i^{*})\}_{i=1}^{|\mathcal{D}|}$, we define robust accuracy as the proportion of test examples that the model answers correctly under every context in $\mathcal{C}$:
\begin{equation}
\mathrm{Acc} = \mathbb{E}_{(x,a^{*}) \sim \mathcal{D}}
\left[\mathbf{1}\left[\forall c \in \mathcal{C},\ f(c \oplus x)=a^{*}\right]\right],
\end{equation}

where $f(c \oplus x)$ denotes the model prediction for the rewritten question. The indicator equals one only when all context-augmented versions of $x$ receive the ground-truth answer $a^{*}$.

%% file: content/text_whitebox_table.tex
\newcommand{\acc}[1]{{\fontsize{10pt}{11pt}\selectfont #1}}
\newcommand{\accstd}[1]{{\textcolor{black!80}{\fontsize{7pt}{8.5pt}\selectfont$\,\pm\,#1$}}}
\newcommand{\accdrop}[1]{\raisebox{-0.45ex}{\textcolor{red}{\fontsize{6pt}{7.5pt}\selectfont$\downarrow #1$}}}
\newcommand{\dropheat}[1]{%
  \pgfmathtruncatemacro{\heatlevel}{3 + 0.75 * #1}%
  \edef\applyheat{\noexpand\cellcolor{red!\heatlevel}}%
  \applyheat%
}
\newcommand{\heatacc}[2]{%
  \dropheat{#2}%
  \acc{#1}%
}
\newcommand{\attackacc}[3]{%
  \dropheat{#3}%
  \acc{#1}\accstd{#2}%
}
\newcommand{\scoreacc}[1]{%
  \pgfmathtruncatemacro{\scorelevel}{5 + 0.65 * #1}%
  \edef\applyscore{\noexpand\cellcolor{red!\scorelevel}}%
  \applyscore%
  \acc{#1}%
}
\newcommand{\compactscoreacc}[1]{%
  \pgfmathtruncatemacro{\scorelevel}{5 + 0.65 * #1}%
  \edef\applyscore{\noexpand\cellcolor{red!\scorelevel}}%
  \applyscore%
  {\fontsize{8pt}{9pt}\selectfont #1}%
}

\begin{table*}[!t]
\centering
\setlength{\tabcolsep}{1.5pt}
\begin{tabular}{ccccccccccc}
\toprule
& \multicolumn{1}{c}{}
& \multicolumn{3}{c}{GSM8K}
& \multicolumn{3}{c}{SVAMP}
& \multicolumn{3}{c}{MultiArith} \\
\cmidrule(lr){1-2}
\cmidrule(lr){3-5}
\cmidrule(lr){6-8}
\cmidrule(lr){9-11}
type & model & Clean & W-5 & W-3 & Clean & W-5 & W-3 & Clean & W-5 & W-3\\
\midrule
\multirow{2}{*}{explicit}
& Llama3.2 & \acc{40.67} & \attackacc{25.33}{5.55}{15.3} & \attackacc{30.22}{0.39}{10.4} & \acc{58.00} & \attackacc{37.00}{3.61}{21.0} & \attackacc{42.67}{2.52}{15.3} & \acc{61.00} & \attackacc{42.67}{2.31}{18.3} & \attackacc{48.33}{2.08}{12.7}\\
& DeepSeek & \acc{77.33} & \attackacc{61.44}{3.42}{15.9} & \attackacc{60.67}{1.53}{16.7} & \acc{80.00} & \attackacc{65.00}{3.00}{15.0} & \attackacc{64.33}{3.06}{15.7} & \acc{98.00} & \attackacc{95.00}{1.00}{3.0} & \attackacc{97.33}{0.58}{0.7}\\
\midrule
\multirow{5}{*}{implicit}
& CODI & \acc{56.00} & \attackacc{3.89}{0.38}{52.1} & \attackacc{7.78}{0.77}{48.2} & \acc{61.00} & \attackacc{2.00}{2.00}{59.0} & \attackacc{5.67}{2.08}{55.3} & \acc{98.00} & \attackacc{7.00}{2.65}{91.0} & \attackacc{14.67}{3.51}{83.3} \\
& SIM-COT & \acc{42.67} & \attackacc{0.89}{0.19}{41.8} & \attackacc{1.00}{0.00}{41.7} & \acc{36.00} & \attackacc{1.67}{2.08}{34.3} & \attackacc{1.00}{1.00}{35.0} & \acc{93.00} & \attackacc{1.33}{0.58}{91.7} & \attackacc{2.33}{2.31}{90.7} \\
& PCCoT & \acc{48.33} & \attackacc{1.11}{0.51}{47.2} & \attackacc{2.67}{0.34}{45.7} & \acc{49.00} & \attackacc{0.67}{0.58}{48.3} & \attackacc{0.33}{0.58}{48.7} & \acc{95.00} & \attackacc{3.67}{2.89}{91.3} & \attackacc{6.67}{1.53}{88.3} \\
& CoLaR & \acc{27.33} & \attackacc{0.22}{0.19}{27.1} & \attackacc{0.67}{0.34}{26.7} & \acc{47.00} & \attackacc{0.00}{0.00}{47.0} & \attackacc{1.00}{0.00}{46.0} & \acc{88.00} & \attackacc{1.00}{0.00}{87.0} & \attackacc{2.67}{1.15}{85.3} \\
& RoT & \acc{24.33} & \attackacc{4.22}{0.96}{20.1} & \attackacc{8.00}{2.85}{16.3} & \acc{49.00} & \attackacc{11.33}{1.53}{37.7} & \attackacc{16.33}{3.51}{32.7} & \acc{58.00} & \attackacc{3.67}{0.58}{54.3} & \attackacc{10.33}{1.53}{47.7} \\
\bottomrule
\end{tabular}
\caption{Accuracy (\%) of textual reasoning models under white-box perturbations. W-5 and W-3 denote adversarial prefixes with 5 and 3 tokens, respectively. Perturbed accuracy is reported as the mean $\pm$ sample standard deviation over three runs. Cell shading encodes the absolute decrease from clean accuracy, with darker red indicating a larger decrease.}
\label{tab:text_whitebox}
\vspace{-2mm}
\end{table*}

%% file: content/experiments.tex
\section{Experiments}

We evaluate textual and multi-modal LRMs under the protocols described in Section~3 and then analyze how adversarial perturbations interact with their latent states. We summarize the main findings before presenting the setup and results.

\begin{takeaway}
\textbf{Takeaway 1.}
Textual implicit reasoning models exhibit significantly lower robustness to prefix perturbations compared to discrete reasoning models, with accuracy dropping sharply under minimal input modifications.
(\cref{sec:text_model_results})

\textbf{Takeaway 2.}
Current multi-modal implicit models exhibit limited robustness under pixel-level perturbations, and the resulting perturbed examples can transfer across models, although transferability varies across datasets.
(\cref{sec:multi_modal_model_results})

\textbf{Takeaway 3.}
Adversarial perturbations substantially reduce the cosine similarity of latent tokens in textual reasoning models, while latent tokens in multi-modal models remain comparatively stable under perturbations.
(\cref{sec:similarity_results})

\textbf{Takeaway 4.}
The latent tokens in multi-modal models are functionally redundant; replacing them with zeros has negligible impact on accuracy, suggesting a weak causal link to the answer.
(\cref{sec:reasoning_results})

\end{takeaway}

\subsection{Setup}

\begin{figure*}[htbp]
\centering
\includegraphics[width=1\textwidth]{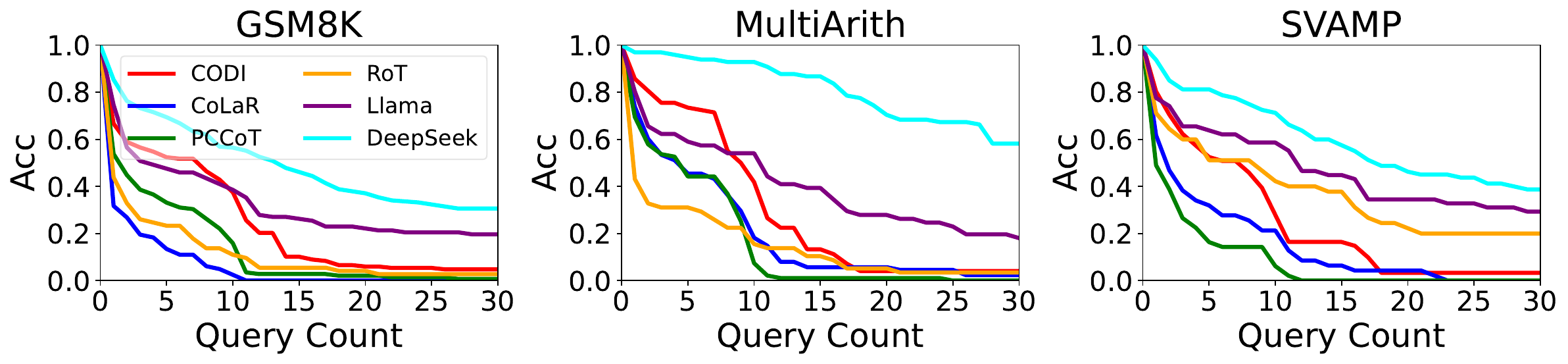}
\caption{Normalized accuracy (\%) of textual reasoning models under black-box perturbations. The x-axis represents the total query count, while the y-axis indicates the normalized accuracy, defined as the ratio of model accuracy with irrelevant information prefixes to the baseline accuracy without prefixes.}
\label{fig:black}
\vspace{-2mm}
\end{figure*}

\begin{table*}[htbp]
\centering
\setlength{\tabcolsep}{1.5pt}
\begin{tabular}{ll *{3}{ccc}}
\toprule
& & \multicolumn{3}{c}{V$^*$}
& \multicolumn{3}{c}{MMVP}
& \multicolumn{3}{c}{MMStar} \\
\cmidrule(lr){1-2}
\cmidrule(lr){3-5}
\cmidrule(lr){6-8}
\cmidrule(lr){9-11}
type & model & Clean & Random & White & Clean & Random & White & Clean & Random & White\\
\midrule
\multirow{1}{*}{explicit}
& Qwen-VL & \acc{42.00} & \heatacc{29.00}{13.0} & \attackacc{8.67}{0.58}{33.3} & \acc{72.67} & \heatacc{54.00}{18.7} & \attackacc{10.89}{2.14}{61.8} & \acc{56.00} & \heatacc{30.00}{26.0} & \attackacc{18.33}{0.58}{37.7}\\
\midrule
\multirow{3}{*}{implicit}
& LVR & \acc{72.00} & \heatacc{64.00}{8.0} & \attackacc{0.33}{0.58}{71.7} & \acc{70.67} & \heatacc{58.67}{12.0} & \attackacc{3.33}{2.40}{67.3} & \acc{62.00} & \heatacc{53.00}{9.0} & \attackacc{11.33}{1.53}{50.7}\\
& Monet & \acc{70.00} & \heatacc{61.00}{9.0} & \attackacc{3.33}{1.53}{66.7} & \acc{68.67} & \heatacc{60.00}{8.7} & \attackacc{3.78}{1.02}{64.9} & \acc{59.00} & \heatacc{51.00}{8.0} & \attackacc{8.67}{1.15}{50.3} \\
& SkiLa & \acc{73.00} & \heatacc{64.00}{9.0} & \attackacc{6.33}{1.15}{66.7} & \acc{68.67} & \heatacc{47.33}{21.3} & \attackacc{1.78}{1.02}{66.9} & \acc{55.00} & \heatacc{35.00}{20.0} & \attackacc{5.00}{1.00}{50.0}\\
\bottomrule
\end{tabular}
\caption{Accuracy (\%) of multi-modal models under perturbations. Random denotes random perturbations, and White denotes perturbations constructed by our white-box method. White-box accuracy is reported as the mean $\pm$ sample standard deviation over three runs. Cell shading encodes the absolute decrease from clean accuracy, with darker red indicating a larger decrease.}%
\label{tab:multi_whitebox}
\vspace{-2mm}
\end{table*}

\begin{table*}[t]
\centering
\begin{tabular}{l *{3}{ccc}}
\toprule
& \multicolumn{3}{c}{V$^*$}%
& \multicolumn{3}{c}{MMVP}%
& \multicolumn{3}{c}{MMStar} \\%
\cmidrule(lr){2-4}%
\cmidrule(lr){5-7}%
\cmidrule(lr){8-10}%
model & LVR & Monet & SkiLa & LVR & Monet & SkiLa & LVR & Monet & SkiLa\\
\midrule
LVR & 0.00 & 38.10 & 6.35 & 2.63 & 97.37 & 17.11 & 17.50 & 47.50 & 30.00\\
Monet & 1.59 & 3.17 & 0.00 & 96.05 & 5.26 & 97.37 & 52.50 & 17.50 & 42.50 \\
SkiLa & 7.94 & 33.33 & 6.35 & 31.58 & 94.74 & 1.32 & 37.50 & 55.00 & 10.00\\
\bottomrule
\end{tabular}
\caption{Transferability of perturbations across multi-modal reasoning models. Rows denote evaluated target models, and columns denote the models used to optimize adversarial perturbations. Accuracy (\%) is reported on questions that all models answer correctly without perturbations.}
\label{tab:transfer}
\vspace{-2mm}
\end{table*}

\begin{table*}[t]
\centering
\begin{tabular}{l *{3}{ccc}}
\toprule
& \multicolumn{3}{c}{GSM8K}%
& \multicolumn{3}{c}{SVAMP}%
& \multicolumn{3}{c}{MultiArith} \\%
\cmidrule(lr){2-4}%
\cmidrule(lr){5-7}%
\cmidrule(lr){8-10}%
model & White & Black & Inter-Q & White & Black & Inter-Q & White & Black & Inter-Q\\
\midrule
CODI & 0.7617 & 0.6312 & 0.2752 & 0.7437 & 0.5057 & 0.3902 & 0.5989 & 0.4766 & 0.4273 \\
SIM-COT & 0.7692 & 0.7174 & 0.4643 & 0.7185 & 0.6640 & 0.4760 & 0.7422 & 0.6564 & 0.5690 \\
PCCoT & 0.7818 & 0.7719 & 0.5563 & 0.5922 & 0.5487 & 0.5148 & 0.7021 & 0.6835 & 0.7262 \\
CoLaR & 0.8942 & 0.8784 & 0.6420 & 0.8421 & 0.8269 & 0.6367 & 0.8320 & 0.8728 & 0.6802 \\
RoT & 0.9381 & 0.9333 & 0.8437 & 0.8817 & 0.8905 & 0.7940 & 0.9147 & 0.9040 & 0.8541 \\
\bottomrule
\end{tabular}
\caption{Latent-token similarity between clean and adversarial prompts for textual reasoning models. White and Black denote adversarial prompts generated under white-box and black-box settings, respectively. Inter-Q denotes the average similarity between latent tokens produced for different questions.}
\label{tab:text_sim}
\end{table*}

\subsubsection{Textual Reasoning Models}
We evaluate five textual implicit reasoning models---CODI~\cite{codi}, SIM-CoT~\cite{simcot}, PCCoT~\cite{pccot}, CoLaR~\cite{colar}, and RoT~\cite{rot}---across three mathematical reasoning datasets: GSM8K~\cite{gsm8k}, MultiArith~\cite{multiarith}, and SVAMP~\cite{svamp}. We randomly sample 100--300 examples from the original test set of each dataset. The robustness evaluation settings include: (1) white-box prefix perturbations with 3-token and 5-token prefixes, (2) black-box prefix perturbations, and (3) random prefix perturbations using 5 random tokens.

For comparison, we additionally evaluate two explicit chain-of-thought reasoning models: \textit{Llama-3.2-1B-Instruct}~\cite{llama}, which serves as the backbone of CODI, and \textit{DeepSeek-R1-Distill-Qwen-1.5B}~\cite{guo2025deepseek}, which serves as the backbone of CoLaR, under the same white-box and black-box prefix perturbation settings.

\subsubsection{Multi-modal Reasoning Models}
For multi-modal implicit reasoning, we test three models---LVR~\cite{lvr}, Monet~\cite{monet}, and SkiLa~\cite{skila}---on V$^*$~\cite{vstar}, MMVP~\cite{mmvp}, and MMStar~\cite{mmstar}, where we randomly sample 100--150 instances from each dataset. For comparison, we also evaluate the explicit reasoning model  Qwen2.5-VL-7B~\cite{qwen25vl}, which serves as the backbone of all three multi-modal implicit reasoning models. We evaluate model robustness under white-box and transfer-based perturbations, with a maximum pixel-level perturbation bound of $\epsilon = 4/255$.

\subsection{Main Results}

\subsubsection{Results on Textual Reasoning Models}
\label{sec:text_model_results}
Table~\ref{tab:text_whitebox} reports the white-box evaluation results on textual reasoning models, where performance degradation on implicit reasoning models is more pronounced than on the explicit reasoning models Llama-3.2-1B-Instruct~\cite{llama} and DeepSeek-R1-Distill-Qwen-1.5B~\cite{guo2025deepseek}. For example, under 5-token prefixes on GSM8K, CODI drops from 56.00\% to 3.89\%, whereas Llama3.2 decreases from 40.67\% to 25.33\%. Figure~\ref{fig:black} presents the results under black-box perturbations, with the corresponding average accuracies reported in Table~\ref{tab:text_blackbox_average} (Appendix~\ref{app:black_box_text_results}), while Figure~\ref{fig:random} (Appendix~\ref{app:random_prefix_text}) presents the results under random prefix perturbations for textual implicit reasoning models. The black-box setting also leads to notable performance degradation.

Notably, the accuracy of implicit reasoning models decreases more rapidly under black-box perturbations than under random prefix perturbations, with a substantial drop observed in early evaluations. In contrast, for the two explicit reasoning models, the degradation trends under black-box and random perturbations are similar, and the final accuracy under black-box perturbations remains significantly higher than that of implicit reasoning models. These results suggest that implicit reasoning models are more sensitive to candidate prefixes containing numerical tokens compared to purely random samples, indicating a higher susceptibility to numerical inputs. This observation is further supported by the white-box results in Table~\ref{tab:number} (Appendix~\ref{app:numerical_prefix_patterns}): under 5-token attacks on GSM8K, numerical tokens occur in 70.06\% of CODI prefixes, compared with only 0.93\% for Llama3.2.

\begin{center}
\begin{minipage}{\columnwidth}
\centering
\small
\setlength{\tabcolsep}{4pt}
\begin{tabular}{lccc}
\toprule
Prefix candidates & Clean & Perturbed & Drop \\
\midrule
Non-numerical only & 56.00 & 26.00 & 30.00 \\
Unrestricted       & 56.00 & 7.78  & 48.22 \\
\bottomrule
\end{tabular}
\captionof{table}{Accuracy (\%) of CODI on GSM8K under optimized prefixes with different candidate-token constraints.}
\label{tab:numerical_token_ablation}
\end{minipage}
\end{center}

To isolate the effect of numerical tokens, we conduct a controlled experiment in which gradient-based prefix optimization is constrained to select only non-numerical tokens. As shown in Table~\ref{tab:numerical_token_ablation}, the clean accuracy of CODI on GSM8K decreases from 56\% to 26\% under the optimized non-numerical prefixes, whereas allowing numerical tokens reduces the accuracy further to 7.78\%. This substantial gap indicates that numerical tokens are an especially important factor in the effectiveness of optimized prefixes.

\subsubsection{Results on Multi-modal Reasoning Models}
\label{sec:multi_modal_model_results}
Table~\ref{tab:multi_whitebox} reports the multi-modal results.  On V$^*$, white-box perturbations reduce LVR from 72.00\% to 0.33\%, compared with a decrease from 42.00\% to 8.67\% for the explicit reasoning model Qwen2.5-VL-7B~\cite{qwen25vl}. This indicates that the vulnerability is not unique to implicit reasoning, although the implicit model undergoes a larger reduction. For random perturbations, the Random column is the proportion of questions answered correctly under all 50 perturbations, while Table~\ref{tab:multimodal_random_average} (Appendix~\ref{app:random_multimodal_results}) reports the average normalized accuracy on clean-correct questions.

Furthermore, we evaluate the cross-model transferability of adversarial examples, as shown in Table~\ref{tab:transfer}. The results vary considerably across datasets and source models: perturbations optimized on SkiLa reduce Monet's accuracy to 0.00\% on V$^*$, but leave it at 97.37\% on MMVP.  The datasets also differ substantially in average image resolution, ranging from $224 \times 224$ for MMVP to $1125 \times 773$ for V$^*$, as reported in Table~\ref{tab:multimodal_resolution} (Appendix~\ref{app:dataset_details}). Because the perturbations are optimized in pixel space, these differences alter the dimensionality and spatial granularity of the optimization and may therefore contribute to the observed variation in transferability.

\begin{table*}[t]
\centering
\small
\setlength{\tabcolsep}{4pt}

\begin{threeparttable}
\begin{tabular*}{\textwidth}{@{\extracolsep{\fill}}ll
    c c c c
}
\toprule
\multirow{2}{*}{Reasoning type}
& \multirow{2}{*}{Model}
& \multicolumn{2}{c}{Clean question}
& \multicolumn{2}{c}{Adversarial question} \\
\cmidrule(lr){3-4}
\cmidrule(lr){5-6}
&
& {\makecell{Clean\\reasoning}}
& {\makecell{Adversarial\\reasoning}}
& {\makecell{Adversarial\\reasoning}}
& {\makecell{Clean\\reasoning}} \\
\midrule

Explicit
& Qwen3
& \scoreacc{70.67} & \scoreacc{35.33} & \scoreacc{32.33} & \scoreacc{70.33} \\

\midrule

\multirow{5}{*}{Implicit}
& CODI
& \scoreacc{56.00} & \scoreacc{33.00} & \scoreacc{2.67} & \scoreacc{25.67} \\
& SIM-COT
& \scoreacc{42.67} & \scoreacc{10.67} & \scoreacc{0.67} & \scoreacc{23.67} \\
& PCCoT
& \scoreacc{48.33} & \scoreacc{25.00} & \scoreacc{0.33} & \scoreacc{30.33} \\
& CoLaR
& \scoreacc{27.33} & \scoreacc{1.00} & \scoreacc{0.00} & \scoreacc{25.33} \\
& RoT
& \scoreacc{24.33} & \scoreacc{23.67} & \scoreacc{0.67} & \scoreacc{1.00} \\

\bottomrule
\end{tabular*}

\end{threeparttable}
\caption{Accuracy (\%) of text reasoning models under latent reasoning-token replacement. Clean question and adversarial question denote the original and adversarial questions, respectively. Clean and adversarial reasoning denote the latent reasoning tokens generated under the corresponding input conditions. Cell shading reflects accuracy, with darker red indicating higher accuracy.}
\label{tab:text_replace}
\end{table*}

\subsection{Analysis of Latent Reasoning}

The results above show that adversarial perturbations can substantially change model outcomes. However, it remains unclear whether these perturbations first disrupt the latent reasoning tokens and thereby alter the final answer, or whether they primarily affect the outcome through mechanisms that bypass latent reasoning. To better understand this distinction, we conduct two additional analyses.

\subsubsection{Do Adversarial Perturbations Affect Latent Tokens?}
\label{sec:similarity_results}

To investigate the impact of adversarial examples on latent tokens, we conduct further experiments. We evaluate the degree of latent-token disruption by computing the cosine similarity between the latent tokens produced for original questions and those produced for adversarial questions. Meanwhile, we compute the mean cosine similarity of latent tokens across different questions as the inter-question reference (Inter-Q). Table~\ref{tab:text_sim} shows substantial variation across textual models. On GSM8K, CODI's similarity falls to 0.6312 under black-box perturbations, whereas RoT remains at 0.9333. CODI's attacked-pair similarity nevertheless remains above its Inter-Q value of 0.2752.

\begin{table}[htbp]
    \centering
    \scriptsize
    \setlength{\tabcolsep}{2pt}
    \resizebox{0.98\columnwidth}{!}{%
    \begin{tabular}{cccccccc}
    \toprule
& \multicolumn{2}{c}{V$^*$}
& \multicolumn{2}{c}{MMVP}
& \multicolumn{2}{c}{MMStar} \\
\cmidrule(lr){2-3}
\cmidrule(lr){4-5}
\cmidrule(lr){6-7}
model & White & Inter-Q & White & Inter-Q & White & Inter-Q \\
    \midrule
        LVR & 0.9397 & 0.8993 & 0.9510 & 0.9081 & 0.9733 & 0.8634 \\
        Monet & 0.9896 & 0.9525 & 0.9861 & 0.9392 & 0.9862 & 0.9336\\
        SkiLa & 0.9839 & 0.9834 & 0.9777 & 0.9795 & 0.9844 & 0.9589\\
    \bottomrule
    \end{tabular}%
    }
    \caption{Latent-token similarity between clean and adversarial inputs for multi-modal models. White denotes adversarial images generated under the white-box setting, while Inter-Q denotes the average similarity between latent tokens produced for different questions.}
    \label{tab:multi_sim}
\end{table}

Table~\ref{tab:multi_sim} presents the evaluation results for multi-modal models, showing that adversarial samples do not cause significant changes in latent tokens. For example, SkiLa's white-box similarity on MMVP is 0.9777, nearly identical to its Inter-Q value of 0.9795, indicating that its latent tokens vary little with either adversarial perturbations or question content.

\subsubsection{Do Latent Tokens Affect the Outcome?}
\label{sec:reasoning_results}

The preceding similarity analysis measures how much latent tokens change under perturbations, but it does not reveal whether these changes influence the final prediction. To examine their functional role, we replace latent tokens between clean and adversarial inputs while keeping the question fixed. Table~\ref{tab:text_replace} reports the results. On clean questions, replacing CoLaR's clean reasoning with adversarial reasoning reduces accuracy from 27.33\% to 1.00\%, whereas RoT changes only from 24.33\% to 23.67\%. This contrast shows that the influence of latent tokens varies substantially across models.

Additionally, we also test an explicit reasoning model, Qwen/Qwen3-0.6B~\cite{yang2025qwen3technicalreport}, whose reasoning tokens can be extracted from the text between \texttt{<think>} and \texttt{</think>}, as a comparison. Replacing its clean reasoning with adversarial reasoning lowers clean-question accuracy from 70.67\% to 35.33\%, indicating that its explicit reasoning tokens strongly influence the outcome.

Table~\ref{tab:multi_replace_compact} presents the results of similar analysis on multi-modal models, showing that replacing latent tokens has little effect on the outcome. On clean V$^*$ questions, LVR remains at 72.00\% whether its latent reasoning is clean, adversarial, or zero. 

This invariance suggests that the final prediction may be largely determined by an answer preference formed before latent reasoning, rather than being revised through the latent tokens. In our white-box assessment, we optimize the logit of a targeted incorrect answer token at the position immediately before latent reasoning begins. The resulting perturbation can therefore directly shift the model's initial answer preference. If the subsequent latent reasoning has little causal influence on the output, it is unlikely to correct this preference, allowing the incorrect judgment to persist in the final answer. By contrast, an explicit reasoning model generates a longer textual reasoning chain, during which it may revisit and correct an initially misleading judgment. This difference may help explain why the multi-modal implicit reasoning models exhibit larger accuracy drops than the explicit baseline under the white-box assessment in \cref{tab:multi_whitebox}.

These experiments show that, in some of the evaluated textual LRMs and all of the evaluated multi-modal LRMs, the latent reasoning chains are largely insensitive to input and have little influence on the final output, consistent with the findings of prior work~\cite{text,mm}.

\begin{table}[htbp]
\centering
\footnotesize
\setlength{\tabcolsep}{1.7pt}
\renewcommand{\arraystretch}{1.1}
\begin{tabular}{l@{\hspace{8pt}}lcccccc}
\toprule
& & \multicolumn{3}{c}{Clean Q} & \multicolumn{3}{c}{Adv Q} \\
\cmidrule(lr){3-5}\cmidrule(lr){6-8}
Dataset & Model
& \makecell{Clean\\R} & \makecell{Adv\\R} & \makecell{Zero\\R}
& \makecell{Adv\\R} & \makecell{Clean\\R} & \makecell{Zero\\R} \\
\midrule
\multirow{4}{*}{V$^*$}
& Qwen-VL  & \compactscoreacc{42.00} & \compactscoreacc{25.00} & \multicolumn{1}{c}{--} & \compactscoreacc{8.00} & \compactscoreacc{25.00} & \multicolumn{1}{c}{--} \\
\cmidrule(lr){2-8}
& LVR   & \compactscoreacc{72.00} & \compactscoreacc{72.00} & \compactscoreacc{72.00} & \compactscoreacc{0.00} & \compactscoreacc{0.00} & \compactscoreacc{0.00} \\
& Monet & \compactscoreacc{70.00} & \compactscoreacc{70.00} & \compactscoreacc{62.00} & \compactscoreacc{2.00} & \compactscoreacc{2.00} & \compactscoreacc{8.00} \\
& SkiLa & \compactscoreacc{73.00} & \compactscoreacc{73.00} & \compactscoreacc{73.00} & \compactscoreacc{5.00} & \compactscoreacc{5.00} & \compactscoreacc{5.00} \\
\midrule
\multirow{4}{*}{MMVP}
& Qwen-VL  & \compactscoreacc{72.67} & \compactscoreacc{31.33} & \multicolumn{1}{c}{--} & \compactscoreacc{9.33} & \compactscoreacc{40.00} & \multicolumn{1}{c}{--} \\
\cmidrule(lr){2-8}
& LVR   & \compactscoreacc{70.67} & \compactscoreacc{70.00} & \compactscoreacc{68.67} & \compactscoreacc{0.67} & \compactscoreacc{5.33} & \compactscoreacc{4.67} \\
& Monet & \compactscoreacc{68.67} & \compactscoreacc{68.67} & \compactscoreacc{60.66} & \compactscoreacc{4.67} & \compactscoreacc{4.67} & \compactscoreacc{10.00} \\
& SkiLa & \compactscoreacc{68.67} & \compactscoreacc{68.67} & \compactscoreacc{68.00} & \compactscoreacc{0.67} & \compactscoreacc{2.67} & \compactscoreacc{4.00} \\
\midrule
\multirow{4}{*}{MMStar}
& Qwen-VL  & \compactscoreacc{56.00} & \compactscoreacc{24.00} & \multicolumn{1}{c}{--} & \compactscoreacc{18.00} & \compactscoreacc{52.00} & \multicolumn{1}{c}{--} \\
\cmidrule(lr){2-8}
& LVR   & \compactscoreacc{62.00} & \compactscoreacc{58.00} & \compactscoreacc{58.00} & \compactscoreacc{11.00} & \compactscoreacc{13.00} & \compactscoreacc{9.00} \\
& Monet & \compactscoreacc{59.00} & \compactscoreacc{59.00} & \compactscoreacc{57.00} & \compactscoreacc{8.00} & \compactscoreacc{8.00} & \compactscoreacc{12.00} \\
& SkiLa & \compactscoreacc{55.00} & \compactscoreacc{50.00} & \compactscoreacc{51.00} & \compactscoreacc{5.00} & \compactscoreacc{7.00} & \compactscoreacc{9.00} \\
\bottomrule
\end{tabular}
\caption{Accuracy (\%) of multi-modal reasoning models with replaced latent tokens. Clean/Adv Q denote clean/adversarial questions; Clean/Adv/Zero R denote clean, adversarial, or zero latent reasoning tokens. Darker red indicates higher accuracy. Qwen-VL does not use zero reasoning tokens.}
\label{tab:multi_replace_compact}
\end{table}

\subsection{Case Study}

We further examine representative cases from textual and multi-modal settings to show how perturbations affect latent reasoning and the final outcome.

Figure~\ref{case1} in Appendix~\ref{app:black_box_text_case} and Figure~\ref{case2} in Appendix~\ref{app:white_box_text_case} illustrate adversarial examples produced under black-box and white-box perturbation settings for textual implicit reasoning models, respectively. They include decoded latent tokens for original and adversarial questions and the final model answers. These examples show how adversarial prefixes can affect the latent reasoning process: some prefix tokens may infiltrate the latent tokens, and prefixes containing large numerical values often induce more numeric latent tokens.  Our experiments show that, in some textual LRMs, the latent reasoning chains exert a measurable influence on the final output. For these models, the disruption of latent reasoning chains by numerical tokens in adversarial prefixes may therefore contribute to the generation of incorrect final answers.

Figure~\ref{case3} (Appendix~\ref{app:multi_modal_case}) presents a multi-modal implicit reasoning example. Although the adversarial image perturbation is nearly imperceptible to human observers, it is sufficient to mislead the model into producing an incorrect answer. The implicit reasoning operations, such as zoom-in and highlight, fail to recover the relevant evidence, suggesting that latent visual reasoning does not improve robustness in this case.

%% file: content/conclusion.tex
Our results reveal a critical tension between the efficiency promised by latent reasoning and its robustness. Across the settings considered in this work, latent reasoning are generally more vulnerable than explicit chain-of-thought, but the source of this vulnerability differs across modalities. For textual LRMs, adversarial prefixes can notably alter latent representations, with numerical cues being particularly disruptive, suggesting that the compressed reasoning dynamics remain sensitive to superficial input patterns. In contrast, the latent states of the evaluated multi-modal models change little under perturbations and replacing or removing them often has only a minor effect on predictions. The result suggests that multi-modal latent tokens could be place-holders that contribute loosely to the final outcome, which makes the model less robust due to the lack of internal thinking process.

More broadly, our findings suggest that compressing explicit reasoning into a small number of continuous states does not automatically preserve the error-correction benefits of a longer reasoning trajectory. Explicit chain-of-thought, despite its higher inference cost, provides additional intermediate computation in which an early misleading signal may be reconsidered or overridden. Therefore, improving latent reasoning should not only be increasing clean-task accuracy or reducing the number of reasoning steps: future methods should also ensure that latent states encode meaningful thinking, exert a genuine causal influence on predictions, and remain stable under task-preserving perturbations. Our study thus highlights adversarial robustness as a necessary dimension, alongside efficiency and accuracy, for evaluating and designing latent reasoning systems.

%% file: content/limitation.tex
Our study has three main limitations. First, the evaluated model set is limited. Although we cover representative text and multi-modal LRMs, current latent reasoning architectures vary substantially in base models, training objectives, and latent-token designs. Therefore, our findings should be viewed as evidence of robustness risks rather than a complete characterization of all LRMs.

Second, our benchmarks and perturbations cover only part of the threat space. We focus on mathematical and visual reasoning tasks, with short textual prefixes for text models and bounded pixel-level noise for multi-modal models. Other settings, such as long-context prompt injection, paraphrase-based attacks, tool-use scenarios, or physically realizable visual perturbations, may reveal different failure modes.

Third, our latent-token analyses are diagnostic rather than fully mechanistic. Cosine similarity and token replacement help probe whether latent states change or affect predictions, but they cannot fully explain how latent reasoning contributes to final answers. More fine-grained causal interventions are needed to understand when latent states are genuinely used and how to make them robust.

%% file: content/appendix.tex
\section{Implementation Details}
\label{app:implementation_details}

\subsection{Text Models}
\paragraph{Model backbones.}
The evaluated textual implicit reasoning models use different backbone language models. CODI is built on \path{meta-llama/Llama-3.2-1B-Instruct}, RoT on Qwen3-VL-2B, and CoLaR on \path{deepseek-ai/DeepSeek-R1-Distill-Qwen-1.5B}. Both PCCoT and SIM-CoT use the 124M-parameter GPT-2 model as their backbone.

For white-box attacks on text implicit reasoning models, we run 30 iterations of optimization, with a candidate prefix pool size of 40 at each iteration.

\paragraph{Modified forward and backward pipeline.}
\Cref{lst:text_whitebox_pipeline} outlines the modified forward and backward propagation pipeline for white-box assessment of textual latent reasoning models. It retains gradients on the input embeddings, unrolls model-specific latent reasoning steps with key--value cache updates, and backpropagates the answer-level objective to the prefix embeddings. Projecting these gradients onto the vocabulary embeddings yields token-level scores for prefix optimization. Model-specific details, such as latent-state transitions and cache handling, are adapted accordingly.

\begin{lstlisting}[
language=Python,
basicstyle=\ttfamily\scriptsize,
breaklines=true,
columns=fullflexible,
keepspaces=true,
showstringspaces=false,
frame=single,
caption={Modified gradient pipeline for the white-box assessment of textual latent reasoning models.},
label={lst:text_whitebox_pipeline}
]
def compute_gradients(input_ids):
    input_embeds = embed_layer(input_ids).requires_grad_(True)

    question_outputs = base_llm(
        inputs_embeds=input_embeds,
        output_hidden_states=True)
    past_kv = question_outputs.past_key_values
    hidden = question_outputs.hidden_states[-1][:, -1:, :]

    for _ in range(num_latent_steps):
        step_outputs = base_llm(
            inputs_embeds=hidden,
            past_key_values=past_kv, ...)
        past_kv = step_outputs.past_key_values
        hidden = step_outputs.hidden_states[-1][:, -1:, :]

    answer_outputs = base_llm(
        inputs_embeds=concat_embeds,
        past_key_values=past_kv)
    answer_token_logits = answer_outputs.logits[answer_pos]

    loss = (answer_token_logits[TARGET]
            - answer_token_logits[BASELINE])
    loss.backward()

    grad = input_embeds.grad
    prefix_grad_emb = grad[:, :prefix_len]
    token_grads = prefix_grad_emb @ embed_weight.T
    return token_grads
\end{lstlisting}

For explicit baselines, we use Llama-3.2-1B-Instruct in the main robustness evaluation because several implicit reasoning models are built on it as the base model, enabling a more directly comparable robustness comparison. In the token-replacement analysis, we instead use Qwen/Qwen3-0.6B because it explicitly generates reasoning traces between \texttt{<think>} and \texttt{</think>}, making its reasoning tokens easy to extract and replace.

For black-box attacks, we construct adversarial prefixes using 30 irrelevant factual statements:
\begin{itemize}
    \itemsep0em
    \item "There are 24 hours in a day."
    \item "April has 30 days."
    \item "A bicycle has 2 wheels."
    \item "Today is the 3rd of the month."
    \item "The current temperature is 20 degrees."
    \item "A year has 12 months."
    \item "A week has 7 days."
    \item "An hour has 60 minutes."
    \item "A minute has 60 seconds."
    \item "A year has 365 days."
    \item "The Earth is about 12742 km in diameter."
    \item "The speed of sound in air is about 343 meters per second."
    \item "Water freezes at 0 degrees."
    \item "The speed of light is about 300000 km/s."
    \item "The human heart has 4 chambers."
    \item "An adult has 32 teeth."
    \item "China has 1.4 billion people."
    \item "Japan has 125 million people."
    \item "The US has 50 states."
    \item "The EU has 27 member states."
    \item "The Nile is about 6650 km long."
    \item "Mount Everest is about 8849 meters high."
    \item "The Pacific covers 165 million km²."
    \item "The human body has 206 bones."
    \item "H2O contains 2 hydrogen atoms."
    \item "Oxygen makes up about 21\% of air."
    \item "Standard atmospheric pressure is 101325 Pa."
    \item "A plant has 46 chromosomes."
    \item "DNA has 2 strands in its double helix."
    \item "Bitcoin was created in 2009."
\end{itemize}

\begin{figure*}[!t]
\centering
\includegraphics[width=1\textwidth]{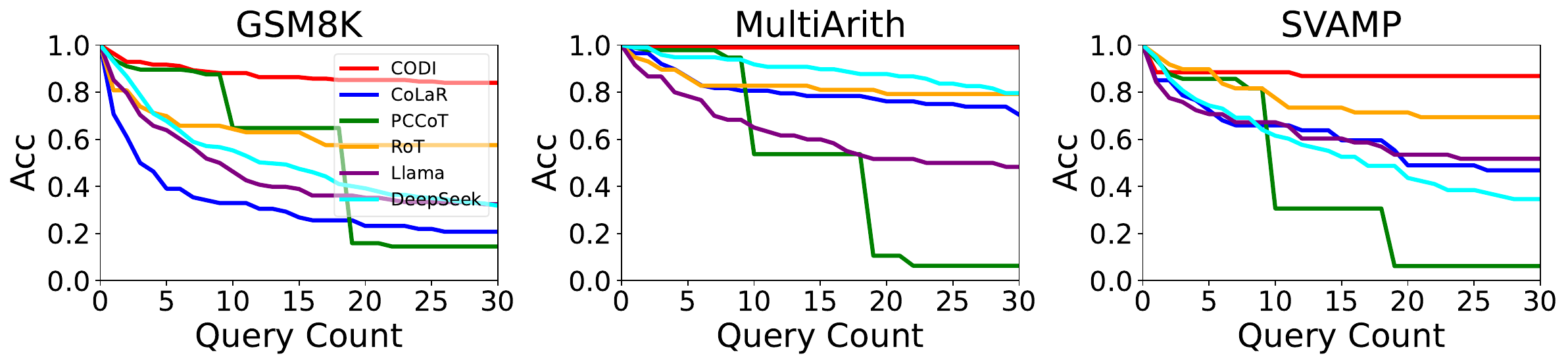}
\caption{Textual reasoning models accuracy (\%) under random perturbation}
\label{fig:random}
\end{figure*}

\subsection{Multi-modal Models}
For white-box attacks on multi-modal models, we run 50 iterations of optimization, along with 50 random perturbations. For each random perturbation, the perturbation at every pixel is sampled independently and uniformly from $\{-4/255,+4/255\}$.

\paragraph{Modified visual preprocessing.}
Because the original Qwen2.5-VL processor contains operations that block end-to-end gradients to the input image, we use functionally equivalent, gradient-compatible resizing, normalization, and patch-merging steps during adversarial optimization, as shown in \cref{lst:modified_processor}. The modified processor is used only to optimize perturbations and propagate gradients to pixel space; evaluation uses the original processor. The implementation is adapted to the tensor layouts and preprocessing configurations of different multi-modal models.

\begin{lstlisting}[
language=Python,
basicstyle=\ttfamily\scriptsize,
breaklines=true,
columns=fullflexible,
keepspaces=true,
showstringspaces=false,
frame=single,
caption={Modified Qwen2.5-VL visual preprocessing steps used during white-box optimization.},
label={lst:modified_processor}
]
def preprocess(img):
    a = F.interpolate(
        img[None],
        size=(H_target, W_target),
        mode='bilinear',
        align_corners=False)[0]
    b = (a - mean) / std
    c, _ = pixel_reshape(b, P, M, T_p)
    return c

def pixel_reshape(x, patch, merge, t):
    n, c, h, w = x.shape
    gt, gh, gw = n // t, h // patch, w // patch
    patches = (x.view(
        gt, t, c, gh // merge, merge, patch,
        gw // merge, merge, patch)
        .permute(0, 3, 6, 4, 7, 2, 1, 5, 8)
        .reshape(gt * gh * gw, c * t * patch * patch))
    return patches, (gt, gh, gw)
\end{lstlisting}

We also attempted a query-based black-box attack that leverages answer probabilities without gradient information, combined with hill climbing search. However, the effectiveness is limited. On the LVR model and MMVP dataset, running 50 queries reduces accuracy from 70.66\% to 57.33\%, and 300 queries further reduce it to 49.33\%. This performance is only marginally better than random perturbation and significantly worse than white-box attacks.

\subsection{Dataset}
\label{app:dataset_details}
Due to the computational cost of attacks, it is impractical to evaluate on the entire dataset. We randomly sample 100 to 300 instances from each dataset: GSM8K (300), MultiArith (100), SVAMP (100), MMVP (150), V$^*$ (100), and MMStar (100). For V$^*$, due to GPU memory constraints, we downsample all images by a factor of two along both height and width to avoid out-of-memory errors during backpropagation.

In the cross-model adversarial-example transferability experiment reported in \cref{tab:transfer}, we evaluate V$^*$, MMVP, and MMStar. Their average image resolutions are reported in \cref{tab:multimodal_resolution}.

\begin{table}[htbp]
\centering
\small
\setlength{\tabcolsep}{7pt}
\begin{tabular}{lccc}
\toprule
Dataset & V$^*$ & MMVP & MMStar \\
\midrule
Resolution & $1125 \times 773$ & $224 \times 224$ & $476 \times 384$ \\
\bottomrule
\end{tabular}
\caption{Average image resolution of the multi-modal datasets.}
\label{tab:multimodal_resolution}
\end{table}

\section{Additional Experiment Results}
\label{app:additional_results}

\subsection{Average Black-box Accuracy of Textual Reasoning Models}
\label{app:black_box_text_results}

Using the notation introduced in \cref{sec:black_box_assessment}, let $\mathcal{D}_{\mathrm{clean}}=\{(x,a^{*})\in\mathcal{D}:f(x)=a^{*}\}$ denote the subset of questions that the model answers correctly without perturbation. We compute this metric over the fixed context pool $\mathcal{C}$ as
\begin{equation}
\begin{aligned}
\mathrm{Acc}_{\mathrm{avg}}
&= \mathbb{E}_{(x,a^{*}) \sim \mathcal{D}_{\mathrm{clean}}}
\left[
\frac{1}{|\mathcal{C}|}
\sum_{c \in \mathcal{C}}
\mathbf{1}\left[f(c \oplus x)=a^{*}\right]
\right].
\end{aligned}
\label{eq:black_box_avg_accuracy}
\end{equation}
Equivalently, for each prefix, we divide the number of questions answered correctly after perturbation by the number answered correctly without perturbation, and then average this ratio over all prefixes. Unlike the robust accuracy in \cref{sec:black_box_assessment}, which requires a question to be answered correctly under every context, this metric measures the average fraction of clean-correct answers retained under perturbation.

\Cref{tab:text_blackbox_average} reports the average normalized accuracy over the 30 fixed irrelevant factual prefixes described in \cref{app:implementation_details}.

\begin{center}
\begin{minipage}{\columnwidth}
\centering
\setlength{\tabcolsep}{4pt}
\renewcommand{\arraystretch}{0.95}
\resizebox{0.98\columnwidth}{!}{%
\begin{tabular}{llccc}
\toprule
Type & Model & GSM8K & SVAMP & MultiArith \\
\midrule
\multirow{2}{*}{explicit}
& Llama3.2 & 0.7260 & 0.7845 & 0.8158 \\
& DeepSeek & 0.8566 & 0.8796 & 0.9711 \\
\midrule
\multirow{5}{*}{implicit}
& CODI    & 0.6389 & 0.6235 & 0.7347 \\
& SIM-CoT & 0.4044 & 0.2778 & 0.4373 \\
& PCCoT   & 0.4444 & 0.4075 & 0.5074 \\
& CoLaR   & 0.3220 & 0.4695 & 0.5864 \\
& RoT     & 0.4694 & 0.7200 & 0.5454 \\
\bottomrule
\end{tabular}
}
\captionof{table}{Average normalized accuracy of textual reasoning models under the black-box assessment, averaged over 30 fixed irrelevant factual prefixes.}
\label{tab:text_blackbox_average}
\end{minipage}
\end{center}

The results show that CoLaR consistently underperforms DeepSeek under black-box perturbations; for example, their average normalized accuracies on GSM8K are 0.3220 and 0.8566, respectively. This indicates that the explicit reasoning model is less sensitive to prefix perturbations than the LRM, consistent with the analysis in \cref{sec:text_model_results}.

\subsection{Average Accuracy under Random Multi-modal Perturbations}
\label{app:random_multimodal_results}
Following the statistical procedure in \cref{app:black_box_text_results}, we restrict the evaluation to questions that each model answers correctly without perturbation. For each of the 50 random pixel-level perturbations described in \cref{app:implementation_details}, we compute the fraction of these questions that remain correctly answered and then average the results across perturbations. \Cref{tab:multimodal_random_average} reports the resulting normalized accuracies following the model and dataset ordering used in \cref{tab:multi_whitebox}.

\begin{table}[htbp]
\centering
\small
\setlength{\tabcolsep}{4pt}
\begin{tabular}{llccc}
\toprule
Type & Model & V$^*$ & MMVP & MMStar \\
\midrule
\multirow{1}{*}{explicit}
& Qwen  & 91.90 & 91.78 & 88.07 \\
\midrule
\multirow{3}{*}{implicit}
& LVR   & 97.17 & 94.15 & 92.74 \\
& Monet & 98.05 & 95.50 & 97.11 \\
& SkiLa & 94.75 & 95.53 & 88.26 \\
\bottomrule
\end{tabular}
\caption{Average normalized accuracy (\%) of multi-modal reasoning models over 50 random pixel-level perturbations.}
\label{tab:multimodal_random_average}
\end{table}

\subsection{Random Prefix Perturbations for Textual Reasoning Models}
\label{app:random_prefix_text}
We evaluate textual implicit reasoning models under random prefix perturbations as a comparison to the black-box prefix search in the main text. Each token in a random prefix is sampled independently and uniformly from the model vocabulary. We prepend these randomly sampled tokens to the input questions and measure the normalized accuracy as the query budget increases. As shown in \cref{fig:random}, random prefixes can reduce accuracy, but the degradation is generally less targeted than the black-box perturbations discussed in \cref{sec:text_model_results}, supporting the observation that numerical or search-selected prefixes are more harmful than purely random tokens.

\subsection{Numerical Patterns in Optimized Textual Prefixes}
\label{app:numerical_prefix_patterns}
We observe that many optimized white-box prefixes contain numbers or numeral-like tokens, so we further quantify how often this pattern occurs. \Cref{tab:number} reports the proportion of such prefixes for 3-token and 5-token attacks. The optimized prefixes for implicit reasoning models frequently contain numerical tokens, while the explicit Llama3.2 baseline rarely exhibits this pattern, indicating that numerical prefix tokens are closely associated with the vulnerability of textual implicit reasoning models.

\begin{table}[htbp]
\centering
\scriptsize
\setlength{\tabcolsep}{1pt}
\noindent\resizebox{0.98\columnwidth}{!}{%
\begin{tabular}{ccccccccccccc}
\toprule
& \multicolumn{1}{c}{}    
& \multicolumn{2}{c}{GSM8K}   
& \multicolumn{2}{c}{SVAMP}  
& \multicolumn{2}{c}{MultiArith} \\ 
\cmidrule(lr){1-2} 
\cmidrule(lr){3-4}  
\cmidrule(lr){5-6} 
\cmidrule(lr){7-8}
type & model & White-5 &  White-3 & White-5 & White-3 & White-5 & White-3\\
\midrule
\multirow{1}{*}{explicit}
& Llama3.2 & 0.0093 & 0.0187 & 0.0179 & 0.0000 & 0.0000 & 0.0000\\
\midrule
\multirow{5}{*}{implicit}
& CODI & 0.7006 & 0.6786 & 0.6721 & 0.6721 & 0.8144 & 0.7755\\
& SIM-COT & 0.5078 & 0.4688 & 0.4722 & 0.3611 & 0.5699 & 0.5699 \\
& PCCoT & 0.8897 & 0.8207 & 0.8571 & 0.8163 & 0.9684 & 0.8526\\
& CoLaR & 0.3780 & 0.2439 & 0.4043 & 0.4043 & 0.5454 & 0.6136\\
& RoT & 0.8767 & 0.7534 & 0.9184 & 0.8367 & 0.8966 & 0.8103\\

\bottomrule
\end{tabular}%
}
\caption{Proportion of prefixes containing numbers}
\label{tab:number}
\end{table}

\begingroup
\setcounter{table}{13}
\begin{table*}[!t]
\centering
\scriptsize
\setlength{\tabcolsep}{6pt}
\resizebox{\textwidth}{!}{%
\begin{tabular}{lccc}
\toprule
Prefix & Latent 1 & Latent 2 & Latent 3 \\
\midrule
None
& \makecell{``4'', ``\textless\textless'', ``:''}
& \makecell{``4'', ``.'', ``-''}
& \makecell{``.'', ``\textgreater\textgreater'', ``+''} \\
\makecell[l]{Non-numerical: \\ \texttt{getic hiringNine}}
& \makecell{``4'', ``\textless\textless'', ``:''}
& \makecell{``4'', ``-'', ``+''}
& \makecell{``.'', ``\textgreater\textgreater'', ``,''} \\
\makecell[l]{Numerical: \\ \texttt{9334000}}
& \makecell{``467'', ``466'', ``477''}
& \makecell{``000'', ``400'', ``4''}
& \makecell{``4'', ``0'', ``004''} \\
\bottomrule
\end{tabular}%
}
\caption{Top-3 decoded tokens at three latent positions under different prefix conditions.}
\label{tab:numerical_latent_case}
\end{table*}
\endgroup
\setcounter{table}{12}

\subsection{Semantic Irrelevance of Optimized Textual Prefixes}
\label{app:prefix_semantic_irrelevance}
\begin{table}[htbp]
\centering
\footnotesize
\setlength{\tabcolsep}{3pt}
\resizebox{\columnwidth}{!}{%
\begin{tabular}{lccc}
\toprule
Method & SVAMP & GSM8K & MultiArith \\
\midrule
CODI    & 0.00\% (0/61) & 0.60\% (1/168) & 0.00\% (0/98) \\
RoT     & 0.00\% (0/49) & 1.37\% (1/73)  & 1.72\% (1/58) \\
PCCoT   & 2.04\% (1/49) & 1.38\% (2/145) & 0.00\% (0/95) \\
CoLaR   & 0.00\% (0/47) & 0.00\% (0/82)  & 0.00\% (0/88) \\
SIM-CoT & 0.00\% (0/36) & 0.00\% (0/128) & 2.15\% (2/93) \\
\bottomrule
\end{tabular}%
}
\caption{Fraction of optimized 3-token prefixes judged as potentially changing the answer to the original question. Parentheses show the number judged answer-changing over the total number evaluated.}
\label{tab:prefix_semantic_check}
\end{table}

Because the white-box assessment optimizes adversarial prefixes directly, we additionally verify that the resulting prefixes do not simply alter the semantics or correct answers of the original problems. We use Qwen3.5-4B as an independent judge and ask it whether each optimized prefix could change the answer to the corresponding original question.

Table~\ref{tab:prefix_semantic_check} reports the results for 3-token prefixes. Across all models and datasets, only a small fraction of optimized prefixes are judged as potentially answer-changing. For 5-token prefixes, none of the optimized prefixes are judged as changing the answer.

These results show that answer-changing prefixes are rare. Therefore, the large accuracy drops under white-box perturbations cannot be explained by genuine semantic changes to the mathematical problems. Instead, they indicate that textual implicit reasoning models are highly sensitive to semantically irrelevant adversarial prefixes.

\subsection{Effect of Numerical Prefixes on Latent Reasoning}
\label{app:numerical_latent_effect}
To isolate the effect of numerical tokens on latent reasoning, we examine a representative CODI example from GSM8K under three conditions: no prefix, an optimized non-numerical prefix, and an optimized numerical prefix. The question asks: \emph{``Nancy is returning her overdue books to the library. She owes \$0.50 cents each on 8 books \ldots{} How much does she have to pay total?''} \Cref{tab:numerical_latent_case} reports the three tokens with the highest decoding probabilities at each of the first three latent positions. The no-prefix and non-numerical-prefix conditions yield highly similar decoded tokens, whereas the numerical prefix introduces multiple numerical tokens. This contrast suggests that the changes in latent reasoning are specifically associated with numerical tokens rather than with the presence of an optimized prefix alone.

\FloatBarrier

\subsection{Black-box Textual Case Study}
\label{app:black_box_text_case}
\begin{figure*}[p]
\centering
\includegraphics[width=1\textwidth]{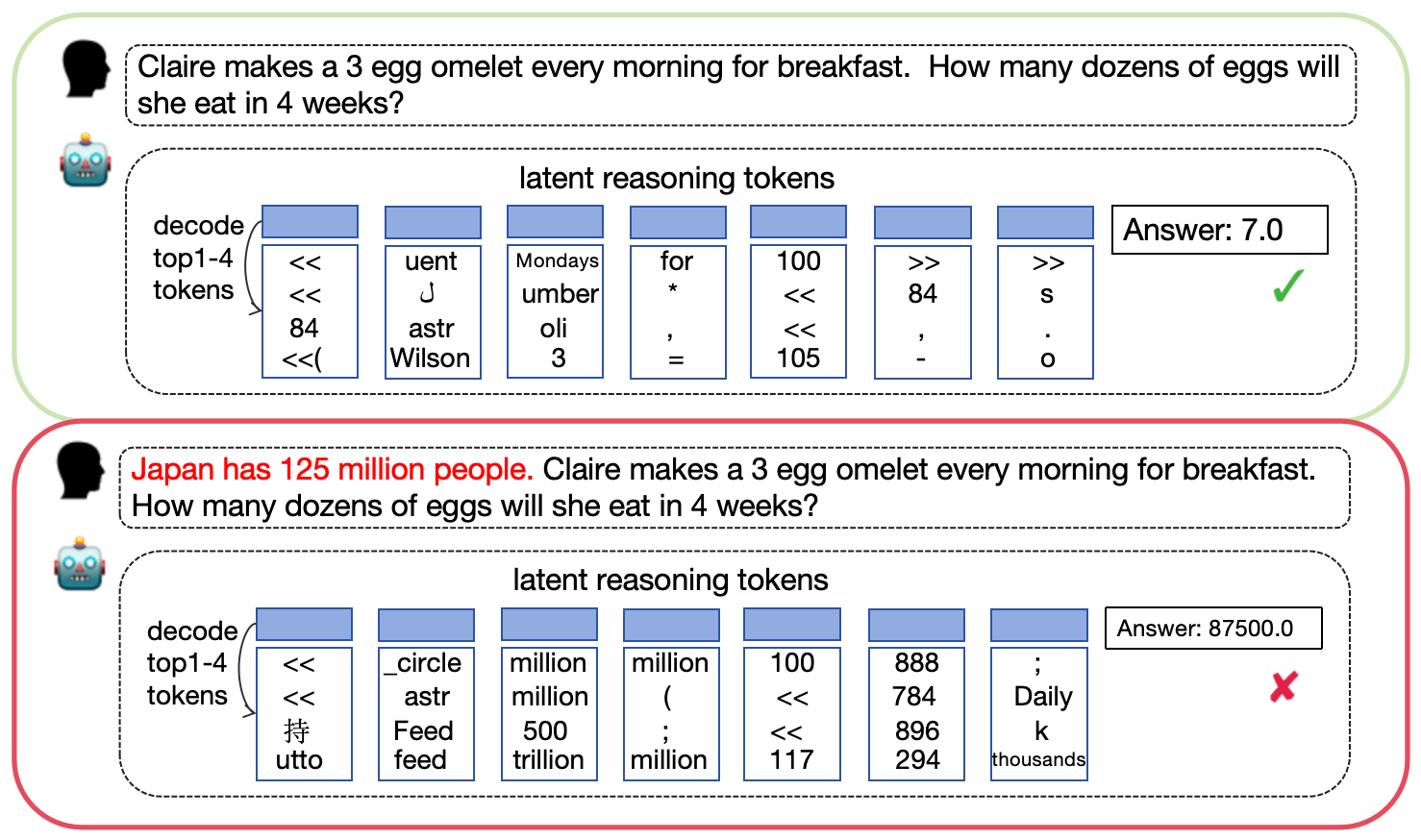}
\caption{Case study 1}
\label{case1}
\end{figure*}

We provide a representative black-box textual example to illustrate how an adversarial prefix changes the model's latent reasoning and answer. As shown in \cref{case1}, the added prefix can influence the decoded latent tokens and steer the model toward an incorrect final response, consistent with the black-box degradation reported in the main results.

\subsection{White-box Textual Case Study}
\label{app:white_box_text_case}
\begin{figure*}[p]
\centering
\includegraphics[width=1\textwidth]{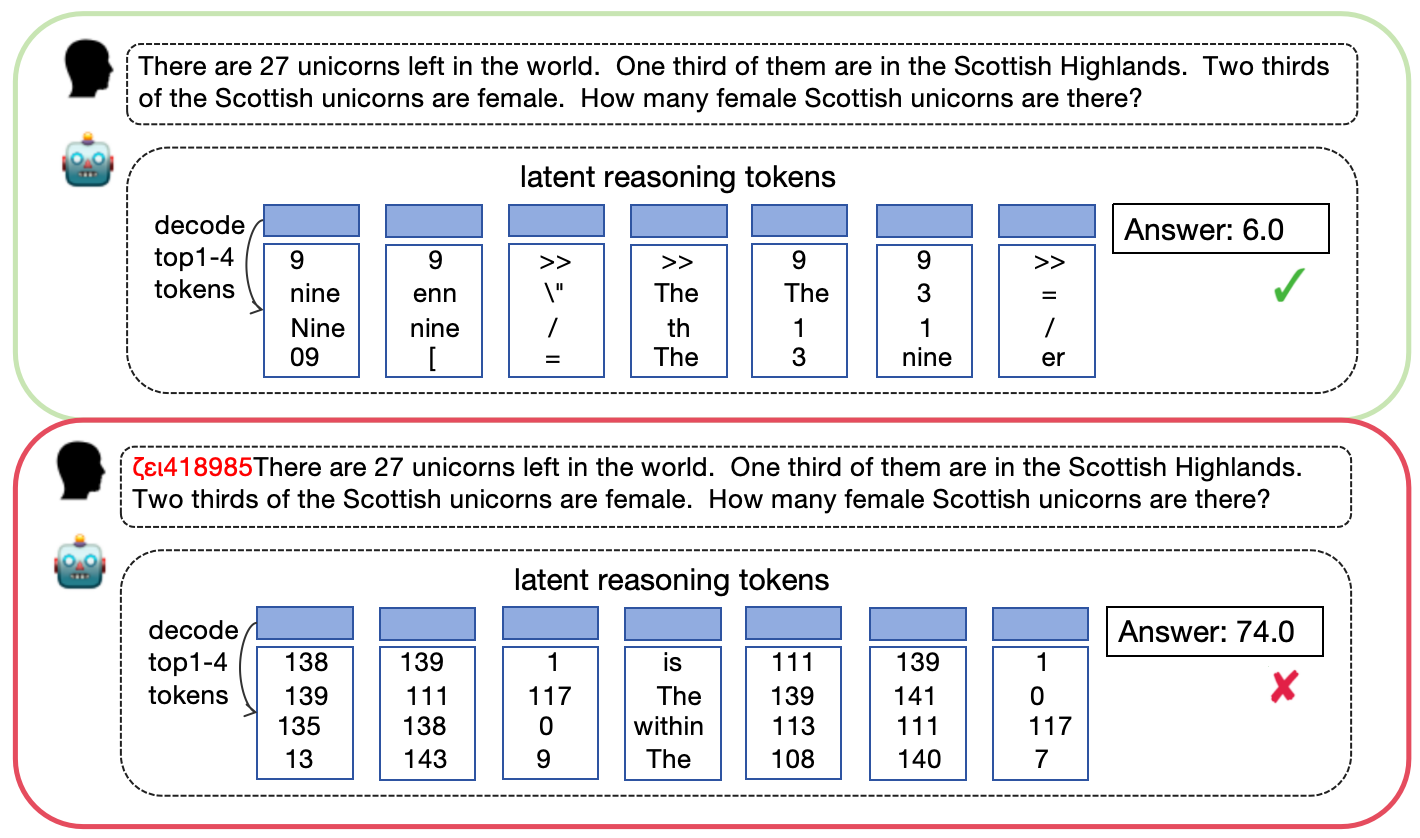}
\caption{Case study 2}
\label{case2}
\end{figure*}

We also include a white-box textual example to show the effect of optimized prefix perturbations. \Cref{case2} visualizes the original and adversarial questions together with their decoded latent tokens and answers. The example shows that a short optimized prefix can alter the latent reasoning trajectory and produce an incorrect answer, providing qualitative evidence for the large white-box accuracy drops in \cref{tab:text_whitebox}.

\subsection{Multi-modal Case Study}
\label{app:multi_modal_case}
\begin{figure*}[p]
\centering
\includegraphics[width=1\textwidth]{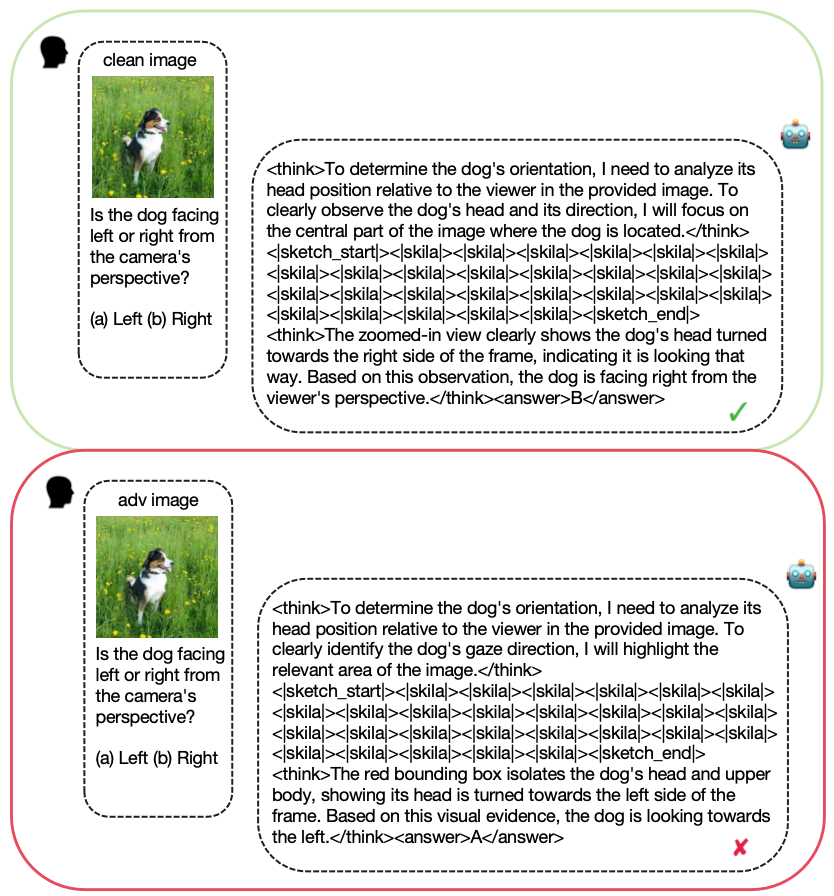}
\caption{Case study 3}
\label{case3}
\end{figure*}

Finally, we present a multi-modal example to illustrate the effect of imperceptible image perturbations. As shown in \cref{case3}, the adversarial image leads the model to an incorrect answer even though the perturbation is visually subtle. The implicit visual reasoning operations do not recover the relevant evidence, matching the quantitative finding that multi-modal latent tokens are stable but weakly tied to the final outcome.

%% file: main.bib
@inproceedings{chain_of_thought,
  author = {Jason Wei and Xuezhi Wang and Dale Schuurmans and Maarten Bosma and Brian Ichter and Fei Xia and Ed Chi and Quoc Le and Denny Zhou},
  title = {Chain-of-Thought Prompting Elicits Reasoning in Large Language Models},
  booktitle = {Advances in Neural Information Processing Systems},
  year = {2022}
}

@article{coconut,
  author = {Shibo Hao and Sainbayar Sukhbaatar and DiJia Su and Xian Li and Zhiting Hu and Jason Weston and Yuandong Tian},
  title = {Training Large Language Models to Reason in a Continuous Latent Space},
  journal = {arXiv preprint ArXiv:2412.06769},
  year = {2024}
}

@article{codi,
  author = {Zhenyi Shen and Hanqi Yan and Linhai Zhang and Zhanghao Hu and Yali Du and Yulan He},
  title = {CODI: Compressing Chain-of-Thought into Continuous Space via Self-Distillation},
  journal = {arXiv preprint ArXiv:2502.21074},
  year = {2025}
}

@article{simcot,
  author = {Xilin Wei and Xiaoran Liu and Yuhang Zang and Xiaoyi Dong and Yuhang Cao and Jiaqi Wang and Xipeng Qiu and Dahua Lin},
  title = {SIM-CoT: Supervised Implicit Chain-of-Thought},
  journal = {arXiv preprint ArXiv:2509.20317},
  year = {2025}
}

@article{pccot,
  author = {Haoyi Wu and Zhihao Teng and Kewei Tu},
  title = {Parallel Continuous Chain-of-Thought with Jacobi Iteration},
  journal = {arXiv preprint ArXiv:2506.18582},
  year = {2025}
}

@article{colar,
  author = {Wenhui Tan and Jiaze Li and Jianzhong Ju and Zhenbo Luo and Ruihua Song and Jian Luan},
  title = {Think Silently, Think Fast: Dynamic Latent Compression of LLM Reasoning Chains},
  journal = {arXiv preprint ArXiv:2505.16552},
  year = {2025}
}

@article{rot,
  author = {Yifan Wang and Shiyu Li and Peiming Li and Xiaochen Yang and Yang Tang and Zheng Wei},
  title = {Render-of-Thought: Rendering Textual Chain-of-Thought as Images for Visual Latent Reasoning},
  journal = {arXiv preprint ArXiv:2601.14750},
  year = {2026}
}

@article{llama,
  author = {{Llama Team}},
  title = {The Llama 3 Herd of Models},
  journal = {arXiv preprint ArXiv:2407.21783},
  year = {2024}
}

@article{lvr,
  author = {Bangzheng Li and Ximeng Sun and Jiang Liu and Ze Wang and Jialian Wu and Xiaodong Yu and Hao Chen and Emad Barsoum and Muhao Chen and Zicheng Liu},
  title = {Latent Visual Reasoning},
  journal = {arXiv preprint arXiv:2509.24251},
  year = {2025}
}

@article{monet,
  author = {Qixun Wang and Yang Shi and Yifei Wang and Yuanxing Zhang and Pengfei Wan and Kun Gai and Xianghua Ying and Yisen Wang},
  title = {Monet: Reasoning in Latent Visual Space Beyond Images and Language},
  journal = {arXiv preprint arXiv:2511.21395},
  year = {2025}
}

@article{skila,
  author = {Jintao Tong and Jiaqi Gu and Yujing Lou and Lubin Fan and Yixiong Zou and Yue Wu and Jieping Ye and Ruixuan Li},
  title = {Sketch-in-Latents: Eliciting Unified Reasoning in MLLMs},
  journal = {arXiv preprint arXiv:2512.16584},
  year = {2025}
}

@article{qwen25vl,
  author = {Shuai Bai and Keqin Chen and Xuejing Liu and Jialin Wang and Wenbin Ge and Sibo Song and Kai Dang and Peng Wang and Shijie Wang and Jun Tang and Humen Zhong and Yuanzhi Zhu and Mingkun Yang and Zhaohai Li and Jianqiang Wan and Pengfei Wang and Wei Ding and Zheren Fu and Yiheng Xu and Jiabo Ye and Xi Zhang and Tianbao Xie and Zesen Cheng and Hang Zhang and Zhibo Yang and Haiyang Xu and Junyang Lin},
  title = {Qwen2.5-VL Technical Report},
  journal = {arXiv preprint arXiv:2502.13923},
  year = {2025}
}

@article{gcg,
  author = {Andy Zou and Zifan Wang and Nicholas Carlini and Milad Nasr and J. Zico Kolter and Matt Fredrikson},
  title = {Universal and Transferable Adversarial Attacks on Aligned Language Models},
  journal = {arXiv preprint ArXiv:2307.15043},
  year = {2023}
}

@article{pgd,
  author = {Aleksander Mądry and Aleksandar Makelov and Ludwig Schmidt and Dimitris Tsipras and Adrian Vladu},
  title = {Towards Deep Learning Models Resistant to Adversarial Attacks},
  journal = {arXiv preprint ArXiv:1706.06083},
  year = {2017}
}

@article{gsm8k,
  author = {Karl Cobbe and Vineet Kosaraju and Mohammad Bavarian and Mark Chen and Heewoo Jun and Lukasz Kaiser and Matthias Plappert and Jerry Tworek and Jacob Hilton and Reiichiro Nakano and Christopher Hesse and John Schulman},
  title = {Training Verifiers to Solve Math Word Problems},
  journal = {arXiv preprint ArXiv:2110.14168},
  year = {2021}
}

@article{svamp,
  author = {Arkil Patel and Satwik Bhattamishra and Navin Goyal},
  title = {Are NLP Models really able to Solve Simple Math Word Problems?},
  journal = {arXiv preprint ArXiv:2103.07191},
  year = {2021}
}

@article{multiarith,
  author = {Subhro Roy and Dan Roth},
  title = {Solving General Arithmetic Word Problems},
  journal = {arXiv preprint ArXiv:1608.01413},
  year = {2016}
}

@article{vstar,
  author = {Penghao Wu and Saining Xie},
  title = {V*: Guided Visual Search as a Core Mechanism in Multimodal LLMs},
  journal = {arXiv preprint ArXiv:2312.14135},
  year = {2023}
}

@article{mmvp,
  author = {Shengbang Tong and Zhuang Liu and Yuexiang Zhai and Yi Ma and Yann LeCun and Saining Xie},
  title = {Eyes Wide Shut? Exploring the Visual Shortcomings of Multimodal LLMs},
  journal = {arXiv preprint ArXiv:2401.06209},
  year = {2024}
}

@article{mmstar,
  author = {Lin Chen and Jinsong Li and Xiaoyi Dong and Pan Zhang and Yuhang Zang and Zehui Chen and Haodong Duan and Jiaqi Wang and Yu Qiao and Dahua Lin and Feng Zhao},
  title = {Are We on the Right Way for Evaluating Large Vision-Language Models?},
  journal = {arXiv preprint ArXiv:2403.20330},
  year = {2024}
}

@article{text,
  author = {Yuyi Zhang and Boyu Tang and Tianjie Ju and Sufeng Duan and Gongshen Liu},
  title = {Do Latent Tokens Think? A Causal and Adversarial Analysis of Chain-of-Continuous-Thought},
  journal = {arXiv preprint ArXiv:2512.21711},
  year = {2025}
}

@article{mm,
  author = {You Li and Chi Chen and Yanghao Li and Fanhu Zeng and Kaiyu Huang and Jinan Xu and Maosong Sun},
  title = {Imagination Helps Visual Reasoning, But Not Yet in Latent Space},
  journal = {arXiv preprint ArXiv:2602.22766},
  year = {2026}
}

@article{seca,
  author = {Buyun Liang and Liangzu Peng and Jinqi Luo and Darshan Thaker and Kwan Ho Ryan Chan and René Vidal},
  title = {SECA: Semantically Equivalent and Coherent Attacks for Eliciting LLM Hallucinations},
  journal = {arXiv preprint ArXiv:2510.04398},
  year = {2025}
}

@article{prompt_var,
  author = {Giannis Chatziveroglou and Richard Yun and Maura Kelleher},
  title = {Exploring LLM Reasoning Through Controlled Prompt Variations},
  journal = {arXiv preprint ArXiv:2504.02111},
  year = {2025}
}

@article{vlm_adv,
  author = {Yunqing Zhao and Tianyu Pang and Chao Du and Xiao Yang and Chongxuan Li and Ngai-Man Cheung and Min Lin},
  title = {On Evaluating Adversarial Robustness of Large Vision-Language Models},
  journal = {arXiv preprint ArXiv:2305.16934},
  year = {2023}
}

@article{vlm_robust,
  author = {Rohit Saxena and Alessandro Suglia and Pasquale Minervini},
  title = {VLM-RobustBench: A Comprehensive Benchmark for Robustness of Vision-Language Models},
  journal = {arXiv preprint ArXiv:2603.06148},
  year = {2026}
}

@article{llm_pertur,
  author = {Qintong Li and Leyang Cui and Xueliang Zhao and Lingpeng Kong and Wei Bi},
  title = {GSM-PLUS: A Comprehensive Benchmark for Evaluating the Robustness of LLMs as Mathematical Problem Solvers},
  journal = {arXiv preprint ArXiv:2402.19255},
  year = {2024}
}

@article{vlm_pertur,
  author = {Zefeng Wang and Zhen Han and Shuo Chen and Fan Xue and Zifeng Ding and Xun Xiao and Volker Tresp and Philip Torr and Jindong Gu},
  title = {Stop Reasoning! When Multimodal LLM with Chain-of-Thought Reasoning Meets Adversarial Image},
  journal = {arXiv preprint ArXiv:2402.14899},
  year = {2024}
}

@misc{o1,
  title={Learning to reason with LLMs},
  author={OpenAI},
  year={2024},
  url = {https://openai.com/index/learning-to-reason-with-llms/}
}

@article{cuadron2025danger,
  title={The Danger of Overthinking: Examining the Reasoning-Action Dilemma in Agentic Tasks},
  author={Cuadron, Alejandro and Li, Dacheng and Ma, Wenjie and Wang, Xingyao and Wang, Yichuan and Zhuang, Siyuan and Liu, Shu and Schroeder, Luis Gaspar and Xia, Tian and Mao, Huanzhi and Thumiger, Nicholas and Desai, Aditya and Stoica, Ion and Klimovic, Ana and Neubig, Graham and Gonzalez, Joseph E.},
  journal={arXiv preprint arXiv:2502.08235},
  year={2025}
}

@article{guo2025deepseek,
  title={Deepseek-r1: Incentivizing reasoning capability in llms via reinforcement learning},
  author={{DeepSeek-AI}},
  journal={arXiv preprint arXiv:2501.12948},
  year={2025}
}

@article{yang2025qwen3technicalreport,
      title={Qwen3 Technical Report}, 
      author={An Yang and Anfeng Li and Baosong Yang and Beichen Zhang and Binyuan Hui and Bo Zheng and Bowen Yu and Chang Gao and Chengen Huang and Chenxu Lv and Chujie Zheng and Dayiheng Liu and Fan Zhou and Fei Huang and Feng Hu and Hao Ge and Haoran Wei and Huan Lin and Jialong Tang and Jian Yang and Jianhong Tu and Jianwei Zhang and Jianxin Yang and Jiaxi Yang and Jing Zhou and Jingren Zhou and Junyang Lin and Kai Dang and Keqin Bao and Kexin Yang and Le Yu and Lianghao Deng and Mei Li and Mingfeng Xue and Mingze Li and Pei Zhang and Peng Wang and Qin Zhu and Rui Men and Ruize Gao and Shixuan Liu and Shuang Luo and Tianhao Li and Tianyi Tang and Wenbiao Yin and Xingzhang Ren and Xinyu Wang and Xinyu Zhang and Xuancheng Ren and Yang Fan and Yang Su and Yichang Zhang and Yinger Zhang and Yu Wan and Yuqiong Liu and Zekun Wang and Zeyu Cui and Zhenru Zhang and Zhipeng Zhou and Zihan Qiu},
      journal={arXiv preprint arXiv:2505.09388},
      year={2025},
      eprint={arXiv preprint arXiv:2505.09388},
}

@article{xu2025towardreasoningsurvey,
  title={Toward large reasoning models: A survey of reinforced reasoning with large language models},
  author={Xu, Fengli and Hao, Qianyue and Shao, Chenyang and Zong, Zefang and Li, Yu and Wang, Jingwei and Zhang, Yunke and Wang, Jingyi and Lan, Xiaochong and Gong, Jiahui and Ouyang, Tianjian and Meng, Fanjin and Yan, Yuwei and Yang, Qinglong and Song, Yiwen and Ren, Sijian and Hu, Xinyuan and Feng, Jie and Gao, Chen and Li, Yong},
  journal={Patterns},
  year={2025},
}

@article{luo2025large,
  title={Large language model agent: A survey on methodology, applications and challenges},
  author={Luo, Junyu and Zhang, Weizhi and Yuan, Ye and Zhao, Yusheng and Yang, Junwei and Gu, Yiyang and Wu, Bohan and Chen, Binqi and Qiao, Ziyue and Long, Qingqing and Tu, Rongcheng and Luo, Xiao and Ju, Wei and Xiao, Zhiping and Wang, Yifan and Xiao, Meng and Liu, Chenwu and Yuan, Jingyang and Zhang, Shichang and Jin, Yiqiao and Zhang, Fan and Wu, Xian and Zhao, Hanqing and Tao, Dacheng and Yu, Philip S. and Zhang, Ming},
  journal={arXiv preprint arXiv:2503.21460},
  year={2025}
}

@article{zhang2025survey,
  title={A survey on test-time scaling in large language models: What, how, where, and how well?},
  author={Zhang, Qiyuan and Lyu, Fuyuan and Sun, Zexu and Wang, Lei and Zhang, Weixu and Hua, Wenyue and Wu, Haolun and Guo, Zhihan and Wang, Yufei and Muennighoff, Niklas and King, Irwin and Liu, Xue and Ma, Chen},
  journal={arXiv preprint arXiv:2503.24235},
  year={2025}
}

@article{kaplan2020scaling,
  title={Scaling laws for neural language models},
  author={Kaplan, Jared and McCandlish, Sam and Henighan, Tom and Brown, Tom B and Chess, Benjamin and Child, Rewon and Gray, Scott and Radford, Alec and Wu, Jeffrey and Amodei, Dario},
  journal={arXiv preprint arXiv:2001.08361},
  year={2020}
}

@article{jiang2026codegenerationsurvey,
  title={A survey on large language models for code generation},
  author={Jiang, Juyong and Wang, Fan and Shen, Jiasi and Kim, Sungju and Kim, Sunghun},
  journal={ACM Transactions on Software Engineering and Methodology},
  year={2026},
}

@article{gao2026far,
  title={How far are we from optimal reasoning efficiency?},
  author={Gao, Jiaxuan and Yan, Shu and Tan, Qixin and Yang, Lu and Xu, Shusheng and Fu, Wei and Mei, Zhiyu and Lyu, Kaifeng and Wu, Yi},
  journal={NeurIPS},
  year={2025}
}

@article{chen2025reasoning,
  title={Reasoning beyond language: A comprehensive survey on latent chain-of-thought reasoning},
  author={Chen, Xinghao and Zhao, Anhao and Xia, Heming and Lu, Xuan and Wang, Hanlin and Chen, Yanjun and Zhang, Wei and Wang, Jian and Li, Wenjie and Shen, Xiaoyu},
  journal={arXiv preprint arXiv:2505.16782},
  year={2025}
}

@article{goodfellow2014explaining,
  title={Explaining and harnessing adversarial examples},
  author={Goodfellow, Ian J and Shlens, Jonathon and Szegedy, Christian},
  journal={arXiv preprint arXiv:1412.6572},
  year={2014}
}

@inproceedings{gan2024reasoning,
  title={Reasoning robustness of llms to adversarial typographical errors},
  author={Gan, Esther and Zhao, Yiran and Cheng, Liying and Yancan, Mao and Goyal, Anirudh and Kawaguchi, Kenji and Kan, Min-Yen and Shieh, Michael},
  booktitle={EMNLP},
  year={2024}
}

@inproceedings{ho2026format,
  title={Format matters: The robustness of multimodal LLMs in reviewing evidence from tables and charts},
  author={Ho, Xanh and Wu, Yun-Ang and Kumar, Sunisth and Boudin, Florian and Takasu, Atsuhiro and Aizawa, Akiko},
  booktitle={AAAI},
  year={2026}
}

@inproceedings{
guha2026openthoughts,
title={OpenThoughts: Data Recipes for Reasoning Models},
author={Etash Kumar Guha and Ryan Marten and Sedrick Keh and Negin Raoof and Georgios Smyrnis and Hritik Bansal and Marianna Nezhurina and Jean Mercat and Trung Vu and Zayne Rea Sprague and Ashima Suvarna and Benjamin Feuer and Leon Liangyu Chen and Zaid Khan and Eric Frankel and Sachin Grover and Caroline Choi and Niklas Muennighoff and Shiye Su and Wanjia Zhao and John Yang and Shreyas Pimpalgaonkar and Kartik sharma and Charlie Cheng-Jie Ji and Yichuan Deng and Sarah M Pratt and Vivek Ramanujan and Jon Saad-Falcon and Stutee Acharya and Jeffrey Li and Achal Dave and Alon Albalak and Kushal Arora and Blake Wulfe and Chinmay Hegde and Greg Durrett and Sewoong Oh and Mohit Bansal and Saadia Gabriel and Aditya Grover and Kai-Wei Chang and Vaishaal Shankar and Aaron Gokaslan and Mike A Merrill and Tatsunori Hashimoto and Yejin Choi and Jenia Jitsev and Reinhard Heckel and Maheswaran Sathiamoorthy and Alex Dimakis and Ludwig Schmidt},
booktitle={The Fourteenth International Conference on Learning Representations},
year={2026},
}

@article{sui2025stop,
title={Stop Overthinking: A Survey on Efficient Reasoning for Large Language Models},
author={Yang Sui and Yu-Neng Chuang and Guanchu Wang and Jiamu Zhang and Tianyi Zhang and Jiayi Yuan and Hongyi Liu and Andrew Wen and Shaochen Zhong and Na Zou and Hanjie Chen and Xia Hu},
journal={Transactions on Machine Learning Research},
year={2025},
}
